%% file: main_all.tex
\documentclass{article}
\usepackage[main,final]{neurips_2026}
\usepackage{url}
\usepackage[utf8]{inputenc}
\usepackage[T1]{fontenc}
\setcitestyle{numbers,square,comma}
\usepackage{hyperref}
\usepackage{caption}
\usepackage{amssymb}
\usepackage{url}
\usepackage{booktabs}
\usepackage{amsfonts}
\usepackage{nicefrac}
\usepackage{microtype}
\usepackage{amsmath}
\usepackage{xcolor}
\usepackage{graphicx}
\usepackage{algorithm}
\usepackage{algpseudocode}
\usepackage{makecell}
\usepackage{wrapfig}
\usepackage[capitalize,nameinlink,noabbrev]{cleveref}
\graphicspath{{../}{./}}
\usepackage{subcaption}
\usepackage{amsthm}
\usepackage[shortlabels]{enumitem}
\newtheorem{theorem}{Theorem}

\newtheorem{proposition}[theorem]{Proposition}

\theoremstyle{definition}
\newtheorem{assumption}{Assumption}
\newtheorem*{remark}{Remark}
\theoremstyle{plain}

\usepackage{xcolor}
\definecolor{darkblue}{rgb}{0.0,0.0,0.7}
\definecolor{darkred}{rgb}{0.4,0,0.3}
\hypersetup{
    colorlinks=true,
    linkcolor=darkred,
    citecolor=darkblue,
    filecolor=darkblue,
    urlcolor=darkblue
}
\crefformat{equation}{#2Equation~(#1)#3}

\usepackage{tocloft}

\newcommand{\listappendixname}{\Large Appendix}
\newlistof{appendix}{apc}{\listappendixname}

\newcommand{\appendixsection}[1]{%
  \section{#1}%
  \addcontentsline{apc}{appendix}{\protect\numberline{\thesection}#1}%
}

\title{
The Hidden Ratio in Adam: Stable Structure, Compression, and Sign Dynamics
}
\author{%
\begin{minipage}{0.98\textwidth}
\centering
\setlength{\tabcolsep}{4pt}
\begin{tabular}{cccc}
\textbf{Yihe Zhou}\textsuperscript{1} &
\textbf{Tongtian Zhu}\textsuperscript{2} &
\textbf{Yingxiao Huo}\textsuperscript{1} &
\textbf{Satya Prakash Dash}\textsuperscript{1}
\end{tabular}\\[0.45em]
\begin{tabular}{ccc}
\textbf{Can Wang}\textsuperscript{2} &
\textbf{Samuel Kaski}\textsuperscript{3,4,1} &
\textbf{Mingfei Sun}\textsuperscript{1}
\end{tabular}\\[0.70em]
{\small
\textsuperscript{1}Department of Computer Science, The University of Manchester, Manchester, UK\\
\textsuperscript{2}College of Computer Science and Technology, Zhejiang University, Hangzhou, China\\
\textsuperscript{3}ELLIS Institute Finland\\
\textsuperscript{4}Aalto University
}\\[0.25em]
{\footnotesize
\texttt{\{yihe.zhou,yingxiao.huo,satyaprakash.dash\}@postgrad.manchester.ac.uk}\\
\texttt{\{samuel.kaski,mingfei.sun\}@manchester.ac.uk}
\\
\texttt{\{raiden,wcan\}@zju.edu.cn}
}
\end{minipage}%
}
\begin{document}

\maketitle

\begin{abstract}

Adam is the default optimizer for training modern deep neural networks, yet its adaptive behavior remains poorly understood due to the complex interaction between its first- and second-moment exponential moving averages (EMAs). 
We study Adam in the tied-$\beta$ regime, where the two EMA decay rates are equal, and show that its adaptive dynamics can be expressed through a transformed ratio with approximately scale-stable behavior. 
Empirically, this transformed ratio exhibits a stable, heavy-tailed distribution across tasks, model scales, and training stages, in contrast to the variability of raw moment magnitudes.
This empirical stability has both practical and conceptual consequences.
First, we derive a recurrence for the transformed ratio, yielding a reparameterization of Adam that replaces the second moment with a compressible state. 
Leveraging its stable distribution, we show that a fixed 4-bit codebook is sufficient in our experiments to store this state without auxiliary scaling, achieving performance competitive with full-precision Adam. 
Second, the transformed ratio view clarifies Adam's connection to sign-based methods: Adam reduces to sign-based momentum modulated by the transformed ratio, and replacing it with a constant recovers Signum as a limiting case.
This perspective further provides a simple rule for transferring learning rates between the two methods. 
Together, these results suggest that tied-$\beta$ Adam admits a simple and approximately stable ratio structure underlying its adaptive behavior and demonstrate its utility for both analysis and efficient implementation.
Code is available at \url{https://github.com/agent-lab/hidden-ratio-adam}.

\end{abstract}

\section{Introduction}

Adam~\citep{kingma2015adam,loshchilov2019decoupled} is the default optimizer for training deep neural networks, from vision models~\citep{dosovitskiy2021image,liu2021swin} to large language models~\citep{devlin2019bert,brown2020language,touvron2023llama,stiennon2020learning,ouyang2022training,bai2022training}.
Yet, despite its ubiquity, a fundamental question remains unresolved: \textit{what determines Adam's adaptive behavior across scales, tasks, and training regimes?} 
In particular, the interaction between its two exponential moving averages (EMAs) remains difficult to interpret. 
Specifically, Adam maintains EMAs of the gradient mean $m_t$ and second moment $v_t$, controlled by $\beta_1$ and $\beta_2$. 
The update normalizes a momentum-like direction by a scale estimate of recent gradients, producing coordinate-wise step sizes.
While this mechanism is simple, its dynamics depend intricately on gradient scale and noise, obscuring a clean, invariant characterization.

Recent large-scale training increasingly uses closer decay rates, 
e.g., $(0.9, 0.95)$~\citep{groeneveld2024olmo,smollm32025,parmar2024reuse,chen2025simpletransfer},
motivating the \textit{tied-$\beta$} regime $\beta_1=\beta_2$. 
This regime is both practically relevant and structurally simpler,
as Adam admits a clearer interpretation:
its update direction resembles a smoothed sign-based momentum, 
while the adaptive denominator modulates the step magnitude through a noise-to-signal ratio of stochastic gradients~\citep{orvieto2025adamsecret}. Complementary analyses suggest that matching the two decay rates introduces useful structural properties, including robustness to gradient scaling and improved stability under nonstationary noise~\citep{dohare2023policycollapse,fernandezhernandez2026adamscale,cattaneo2026minibatchnoiseadam}.

In this work, we show that tied-$\beta$ Adam admits an unexpected simplification:
its adaptive dynamics can be reparameterized through an approximately scale-stable ratio. 
Specifically, decomposing the second moment and factoring out the shared $\beta$-dependent scale yields a natural-scale ratio $y_t$ (i.e., \textit{the hidden ratio}) that isolates stochastic fluctuations from deterministic scaling. 
Our key finding is that $y_t$ follows a remarkably stable distribution across models, tasks, and training stages. 
Unlike raw moment magnitudes, which vary with gradient scale,
the distribution of $y_t$ is concentrated with a persistent heavy right tail. 
This suggests a simple underlying structure: 
tied-$\beta$ Adam combines a deterministic scale with an approximately stable law governing adaptive attenuation.

This perspective has two immediate consequences.
The first is representational.
In the tied-$\beta$ regime, Adam can be equivalently parameterized by $(m_t, y_t)$ instead of $(m_t, v_t)$. 
Since $y_t$ has a stable and compressible distribution, it admits aggressive quantization. 
We show that a fixed 4-bit codebook (16 levels) suffices to nearly match full-precision Adam performance,
without requiring additional block-wise or tensor-wise scaling factors. 
To our knowledge, this is the first result demonstrating that a transformed second-moment-derived state in Adam can be quantized to 4-bit precision with a fixed codebook while retaining strong performance.
The second consequence is both practical and conceptual.
The ratio formulation clarifies Adam's relationship to Signum (SignSGD with momentum)~\cite{sun2023momentumsignsgd,bernstein2018signsgd}.
In the tied-$\beta$ regime, Adam can be viewed as a sign-based momentum method whose step magnitude is attenuated by $y_t$. 
Averaging this attenuation under the stable distribution of $y_t$ yields a simple prediction for the effective learning-rate ratio between Adam and its sign-dominated counterpart, 
which aligns with our empirical observations.
Furthermore, replacing $y_t$ with a constant reduces Adam to a Signum-type method with fixed attenuation, showing that a constant-state approximation recovers a Signum-like limit of tied-$\beta$ Adam. This view provides a principled way to transfer or initialize learning rates between the two methods.

We do not claim that tied-$\beta$ Adam universally dominates standard Adam configurations or that the transformed-ratio law fully characterizes optimizer dynamics globally. Rather, our contributions are:
\begin{itemize}[leftmargin=*]
\item We identify a scale-stable ratio $y_t$ in tied-$\beta$ Adam with a stable, heavy-tailed distribution across tasks and models. Truncating or coarsely discretizing $y_t$ incurs negligible performance loss.
\item We identify and derive a recursive characterization of $y_t$, enabling a $(m_t,y_t)$ parameterization and a 4-bit fixed-codebook implementation without auxiliary scaling.
\item The ratio view yields a learning-rate transfer rule and recovers Signum as a constant-state limit.
\end{itemize}

\section{Background and Related Work}
\label{sec:background_setup}
\paragraph{Adam~\citep{kingma2015adam} and Signum~\citep{bernstein2018signsgd,sun2023momentumsignsgd}.}
As an iterative adaptive optimizer, \textbf{Adam} maintains exponential moving averages of the gradient $g_t$ and its square at each iteration $t$,
\begin{align}
    m_t := \beta_1 m_{t-1} + (1-\beta_1) g_t, 
    \qquad
    v_t := \beta_2 v_{t-1} + (1-\beta_2) g_t^2.
\end{align}
Since both averages are initialized to zero, 
Adam uses the bias-corrected elementwise scaling:
\begin{align}
    \hat m_t := \frac{m_t}{1-\beta_1^t},
    \qquad
    \hat v_t := \frac{v_t}{1-\beta_2^t},
    \qquad
    \Delta \theta_t
    :=
    -\eta
    \frac{\hat m_t}{\sqrt{\hat v_t}+\epsilon}, 
\end{align}
where $\epsilon$ is a small positive constant for numerical stability. 
Thus, Adam can be viewed as normalizing an exponential estimate of the gradient mean by an exponential estimate of its root second moment.
In contrast, \textbf{Signum} maintains a first-moment EMA and applies a sign update:
\begin{align}
    m_t := \beta m_{t-1} + (1-\beta) g_t,
    \qquad
    \Delta \theta_t
    :=
    -\eta \operatorname{sign}(m_t).
\end{align}
The update keeps only the direction of the momentum and uses a fixed coordinatewise step magnitude.
A growing line of work has sought to understand Adam through structural interpretations of its update rule, rather than only through empirical tuning heuristics. 
Early analyses separated the roles of momentum and adaptive scaling, interpreting Adam as combining a sign-based directional component with a variance-controlled magnitude term~\cite{balles2018dissecting}, or more generally disentangling momentum from adaptivity in Adam-style methods~\cite{ziyin2020laprop}. 
More recent work further expresses Adam's adaptive scaling through a local signal-to-noise ratio, giving the denominator a direct statistical interpretation~\cite{orvieto2025adamsecret}. 
Under this view, Signum~\cite{bernstein2018signsgd,sun2023momentumsignsgd} can be related to steepest descent under an $\ell_\infty$ trust-region geometry, while Adam corresponds to a signed-momentum direction whose effective step magnitude is modulated by local signal-to-noise structure.
Our work follows this structural perspective, but identifies within Adam's scaling term a comparatively stable distributional quantity that can be represented directly.

\paragraph{Tied-\texorpdfstring{$\beta$}{beta} regime.}
Our analysis focuses on the tied-\(\beta\) case, \(\beta_1=\beta_2=\beta\). Ignoring bias correction and \(\epsilon\) for clarity, the adaptive factor reduces to $\frac{m_t}{\sqrt{v_t}}$.
Because \(m_t\) and \(v_t\) are then computed with the same exponential weights, \(v_t-m_t^2\) has a natural variance-like interpretation. In fact, \(v_t\ge m_t^2\) for all gradient sequences and all \(t\) if and only if \(\beta_1=\beta_2\); the proof is deferred to Appendix~\ref{app:balanced_regime_motivation}.
In this case, Adam admits a particularly clean online mean--variance interpretation~\cite{orvieto2025adamsecret}, and satisfies a first-order gradient scale-invariance property that singles out tied $\beta$ values as a structurally meaningful choice~\cite{fernandezhernandez2026adamscale}. 
Analyses of mini-batch noise further suggest that the preferred relation between $\beta_1$ and $\beta_2$ depends on the noise regime, with larger batch sizes often favoring $\beta_1$ closer to $\beta_2$ in terms of validation performance~\cite{cattaneo2026minibatchnoiseadam}. 
Complementary empirical studies in nonstationary optimization and reinforcement learning indicate that mismatched first- and second-moment timescales can weaken normalization and destabilize updates~\cite{ellis2024adamrel,dohare2023policycollapse}, whereas tied or near-tied decay rates can improve stability in practice~\cite{moalla2024norepresentation,lossplasticity2024,goldie2024learnedopt}. 
Motivated by these observations, we focus on $\beta_1=\beta_2$ when analyzing Adam's ratio recursion and distributional behavior, since this is the regime where prior work suggests both cleaner dynamics and improved stability.

\section{Hidden Ratio: Theoretical and Empirical Properties}
In the tied-\(\beta\) regime, define
\begin{align*}
    r_t := \frac{v_t-m_t^2}{m_t^2},
    \quad
    \frac{m_t}{\sqrt{v_t}}
    =
    \frac{\operatorname{sign}(m_t)}{\sqrt{1+r_t}} .
\end{align*}
Thus the update direction is determined by the sign of the momentum, while \(r_t\) controls the magnitude attenuation. Since \(r_t\ge 0\) under tied $\beta$ values, the denominator is bounded below by \(1\)~\citep{orvieto2025adamsecret}.
We now define the \textbf{transformed ratio}
\begin{align}
    y_t := \frac{1-\beta}{\beta} r_t,
    \qquad
    \frac{m_t}{\sqrt{v_t}}
    =
    \frac{\operatorname{sign}(m_t)}
    {\sqrt{1+\frac{\beta}{1-\beta}y_t}} .
    \label{eq:sign-over-ratio}
\end{align}
This reparameterization identifies \(y_t\) as the central quantity in the Adam update, thereby motivating a closer examination of its statistical structure.
Intuitively, $y_t$ measures the variance-like residual in Adam's denominator after removing the deterministic scale induced by tied exponential averaging.

\subsection{Statistical Structure of \texorpdfstring{\(y_t\)}{y t}: Stability and Heavy Tails}
Since mini-batches are sampled independently, we follow the diffusion-approximation view of mini-batch SGD and model the local gradient noise by independent
Gaussian fluctuations~\citep{mandt2017sgd,he2019control,li2022zeroloss}.
\begin{assumption}[Local Gaussian window]
\label{ass:local-gaussian-window}
For one coordinate in a local time window, with \(\sigma>0\),
\begin{equation*}
    g_{t-j}=\mu+\sigma\xi_j,
    \qquad
    \xi_j\stackrel{\mathrm{i.i.d.}}{\sim}\mathcal N(0,1),
    \qquad j\ge0 .
\end{equation*}
\end{assumption}
We use an infinite-history local reference, omitting initialization transients. Fix \(0<\beta<1\) and let
\(w_j=(1-\beta)\beta^j\),
\(m_t=\sum_{j\ge0}w_jg_{t-j}\), \(v_t=\sum_{j\ge0}w_jg_{t-j}^2\),
\(U_t=v_t-m_t^2\), \(y_t=\frac{1-\beta}{\beta}\frac{U_t}{m_t^2}\),
\(s_\beta^2=\sum_{j\ge0}w_j^2=\frac{1-\beta}{1+\beta}\),
\(\sigma_m=\sigma s_\beta\), \(a=\mu/\sigma_m\), and
\(Z=s_\beta^{-1}\sum_{j\ge0}w_j\xi_j\). 
Under Assumption~\ref{ass:local-gaussian-window}, \(Z\sim\mathcal N(0,1)\) and
\(m_t=\sigma_m(Z+a)\). When comparing many local windows, let \(\mathcal I\)
index the observed coordinate-window instances and let \(a_i\) be the local
shift at index \(i\). Treating these instances as equally weighted samples gives
the empirical pooled-shift distribution
\begin{equation*}
    \widehat P_A
    :=
    \frac{1}{|\mathcal I|}\sum_{i\in\mathcal I}\delta_{a_i},
    \qquad
    A\sim\widehat P_A,
    \qquad
    \mathbb E_{\widehat P_A}[A^2]
    =
    \frac{1}{|\mathcal I|}\sum_{i\in\mathcal I}a_i^2 .
\end{equation*}
Equivalently, \(A\sim\widehat P_A\) means drawing \(I\) uniformly from
\(\mathcal I\) and setting \(A=a_I\). More generally, a pooled-shift
distribution \(P\) is any distribution over local shifts used in this sampling
role; the empirical case is \(P=\widehat P_A\). For any fixed scalar shift \(b\),
write \(Y_b:=2/(Z+b)^2\), so a single local window uses \(Y_a\). If \(A\sim P\)
is independent of \(Z\), the pooled denominator reference is
\(Y_A:=2/(Z+A)^2\).

\begin{proposition}[Shifted inverse-square reference for the ratio]
\label[proposition]{thm:y-structural-decomposition}
Under Assumption~\ref{ass:local-gaussian-window} and the notation above, the
transformed ratio satisfies
\begin{equation}
\label{eq:yt-relative-reference}
    y_t
    =
    Y_a\left(1+\frac{\eta_t}{2}\right),
    \qquad
    \eta_t
    =
    \eta_t^{\mathrm{coup}}+\eta_t^{\mathrm{fluc}},
\end{equation}
where \(\gamma_\beta=\frac{1-\beta}{1+\beta+\beta^2}\),
\(\eta_t^{\mathrm{coup}}:=\gamma_\beta(Z^2-1)\), and
\(\eta_t^{\mathrm{fluc}}
    :=\frac{1-\beta}{\beta}
    \frac{U_t-\mathbb E[U_t\mid Z]}{\sigma_m^2}\).
Let \(P\) be a pooled-shift distribution with finite fourth moment, and let
\(A\sim P\) be independent of \(Z\). Then the fixed shifted reference
\begin{equation}
\label{eq:yc-reference}
\colorbox{black!5}{$\displaystyle
\begin{aligned}
    Y_c=\frac{2}{(Z+c)^2},
    \qquad Z\sim\mathcal N(0,1),
    \qquad c^2:=\mathbb E_P[A^2]
\end{aligned}
$}
\end{equation}
matches the mixed denominator reference \(Y_A=2/(Z+A)^2\) through second order:
for every fixed \(k>0\), writing \(q=\sqrt{2/k}\), and denoting the standard
normal CDF and density by \(\Phi\) and \(\phi\),
\begin{align}
    \Pr(Y_A>k)
    &=
    2\Phi(q)-1-q\phi(q)\mathbb E_P[A^2]
    +O(\mathbb E_P[A^4]), \nonumber\\
    \Pr(Y_c>k)
    &=
    2\Phi(q)-1-q\phi(q)c^2+O(c^4).
    \label{eq:second-order-survival-expansion}
\end{align}
In particular, for a family \(P_\varepsilon\) of pooled-shift distributions with
\(A_\varepsilon\sim P_\varepsilon\),
\(c_\varepsilon^2:=\mathbb E_{P_\varepsilon}[A_\varepsilon^2]\), and
\(\mathbb E_{P_\varepsilon}[A_\varepsilon^4]=O(\varepsilon^4)\), the
corresponding references satisfy
\(\Pr(Y_{A_\varepsilon}>k)-\Pr(Y_{c_\varepsilon}>k)=O(\varepsilon^4)\).
\end{proposition}

\begin{remark}
Equation~\ref{eq:yt-relative-reference} is a denominator-reference decomposition:
for a fixed local shift \(a\), \(Y_a=2/(Z+a)^2\) carries the inverse-square
denominator, while \(\eta_t\) keeps the numerator as an exact relative
correction. The pooled-shift distribution \(P\) is the formal version of
flattening many observed local shifts; drawing \(A\sim P\) selects one
representative local shift, so the denominator reference becomes
\(Y_A=2/(Z+A)^2\). Equation~\ref{eq:second-order-survival-expansion} shows that the
survival curve of \(Y_A\) depends to second order only on \(\mathbb E_P[A^2]\).
Choosing \(c=\sqrt{\mathbb E_P[A^2]}\) keeps this leading shift effect and the
same inverse-square singularity, which is why we use \(Y_c\) in~\cref{eq:yc-reference} as the reference for the empirical \(y_t\). 
\end{remark}
\begin{wrapfigure}{r}{0.40\textwidth}
   \vspace{-1.8em}
    \centering
    \includegraphics[width=0.40\textwidth]{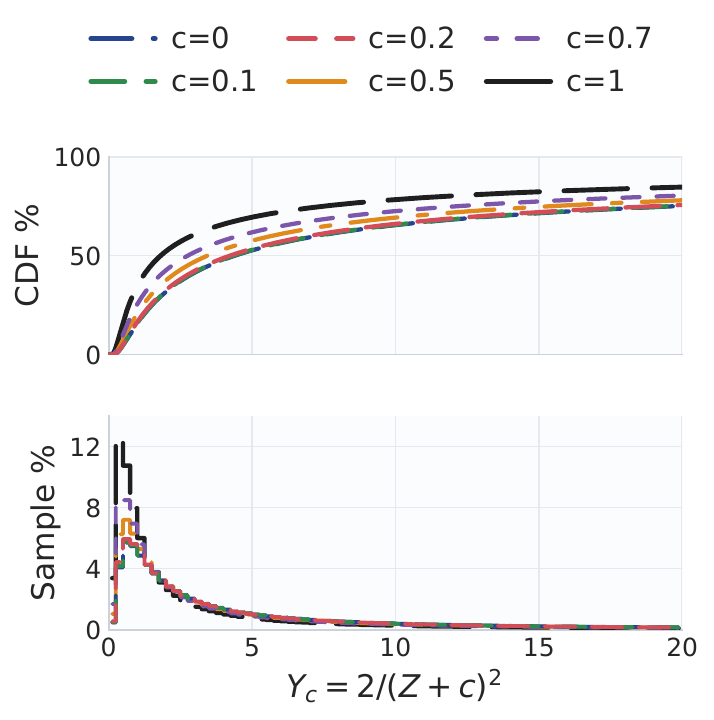}
     \vspace{-1.8em}
    \caption{\small Shifted inverse-square reference \(Y_c=2/(Z+c)^2\). For
    \(c\le0.2\), the CDF and histogram are nearly unchanged; the shape is driven
    mainly by the inverse-square form.}
    \label{fig:shifted-inverse-square}
\end{wrapfigure}

\begin{figure*}[htb]
    \centering
       \vspace{-1.2em}
    \includegraphics[width=.9\textwidth]{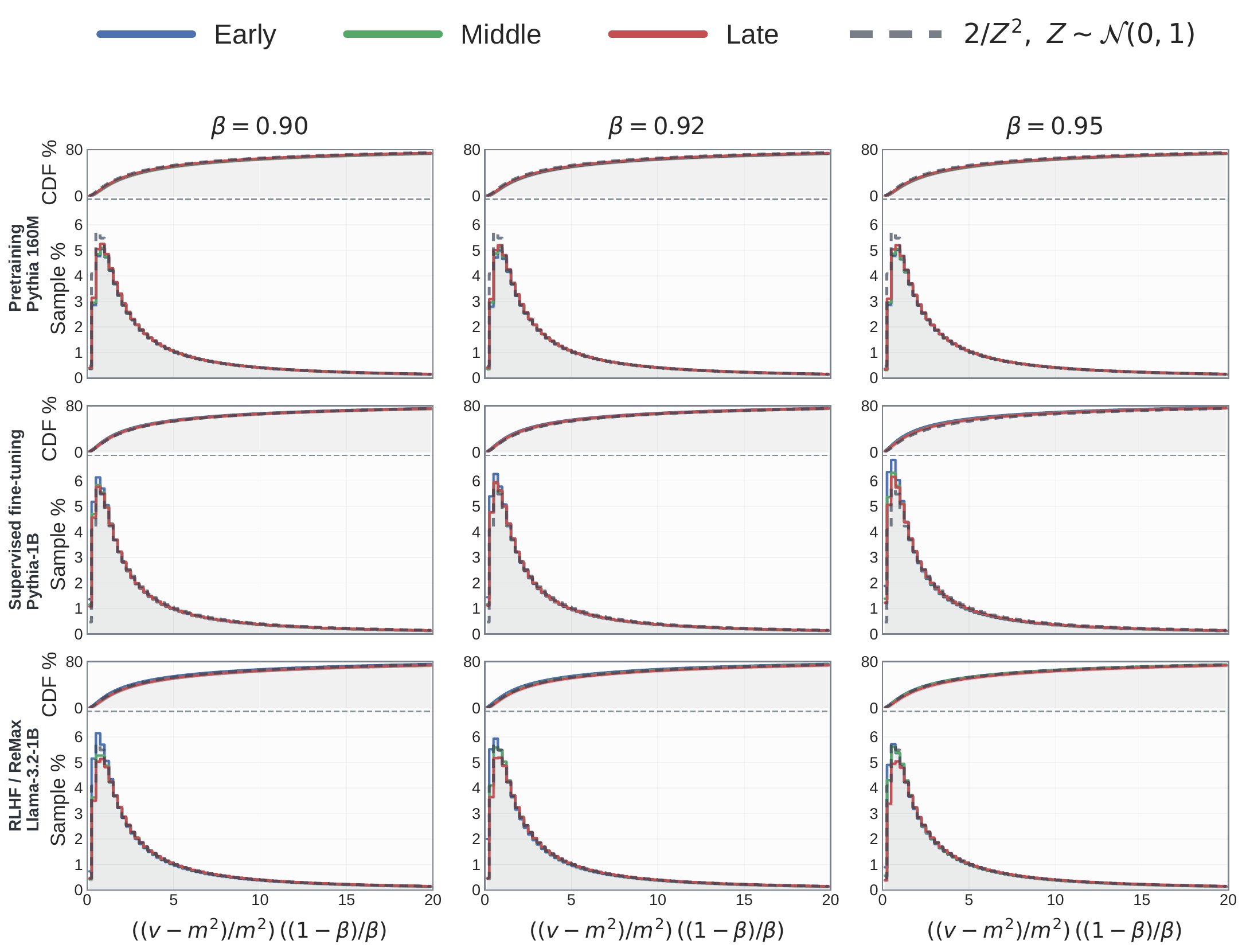}
    \caption{
    Distribution of the transformed ratio $y_t$ across training regimes.
    Columns correspond to \(\beta=0.90, 0.92, 0.95\), while rows compare
    pretraining on Pythia-160M, supervised fine-tuning on Pythia-1B, and
    RLHF on Llama-3.2-1B-Instruct.
    Within each block, the top panel shows the CDF and the bottom panel shows
    the histogram; colors denote early, middle, and late training stages.
    }
    \label{fig:y_dist_pretrain_sft_rlhf}
\end{figure*}

\Cref{fig:shifted-inverse-square} illustrates this point for the reference
distribution. When the shift \(c\) is small, the CDF and histogram of
\(Y_c=2/(Z+c)^2\) remain close to those of the unshifted reference \(2/Z^2\),
preserving both the bulk and the right-tail shape. This suggests that the
inverse-square denominator singularity drives the dominant distributional form,
while local shifts and numerator fluctuations introduce only secondary
distortions. The full derivation is given in Appendix~\ref{app:y-distribution}.
This picture is reflected in the empirical distributions in
\cref{fig:y_dist_pretrain_sft_rlhf}. Across pre-training, SFT, RLHF, and
\(\beta\in\{0.90,0.92,0.95\}\), the transformed ratio \(y_t\) consistently shows
a compact single-digit bulk with a persistent right tail. Within each setting,
the early-, middle-, and late-stage curves remain close, indicating only mild
distributional drift during training. The empirical CDFs also stay close to the
\(2/Z^2\) reference over a broad range. Together, these observations support the
view that the rescaled ratio \(y_t\) captures a natural-scale denominator state
whose distribution is more stable than the raw moment magnitudes \(m_t\) and
\(v_t\).

\subsection{Robustness of \texorpdfstring{\(y_t\)}{y t} to coarse quantization}

Given that the distribution of 
\(y_t\)
 exhibits a clearly concentrated bulk together with a long right tail, we now investigate whether a very coarse quantization of 
\(y_t\)
 is sufficient to retain behavior and performance close to those of the original Adam. To isolate its role, we keep the original Adam recursions for \(m_t\) and \(v_t\) unchanged, compute \(y_t\) from the resulting states, and replace only this variable by a coarse low-precision approximation before reconstructing the adaptive denominator.

 \begin{figure*}[tb]
    \vspace{-1em}
    \centering
    \includegraphics[width=.9\textwidth]{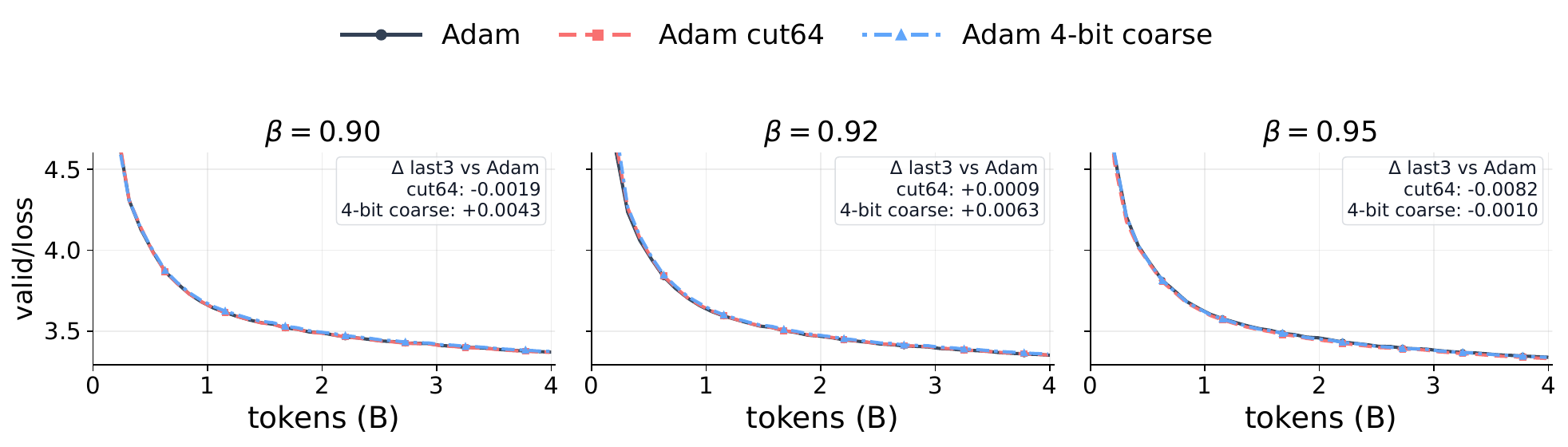}
\caption{
Comparison of the training performance of standard Adam and its modified variants based on the transformed ratio \(y_t\), including long-tail truncation and coarse 4-bit quantization. 
Even after truncating the right tail and representing the remaining bulk with only 16 quantization levels, the resulting optimizer still closely tracks the behavior and final performance of the standard Adam.
}
    \label{fig:y_4bit_result}
      \vspace{-1.0em}
\end{figure*}
We begin with a particularly simple modification: truncating the upper tail of \(y_t\). 
The cut64 variant replaces \(y_t\) by \(\min(y_t,64)\) before reconstructing the adaptive denominator, thereby removing only rare far-tail values. 
Empirically, this truncation has little effect, and the resulting optimizer still tracks standard Adam closely.

We then take a more aggressive step and replace \(y_t\) itself by a coarse low-precision approximation. 
After the same cut64 truncation, we quantize \(y_t\) using the fixed 4-bit codebook
\(\mathcal C=\{0.25,0.5,0.75,1,1.5,2,3,4,6,8,12,16,24,32,48,64\}\),
an FP4-like non-uniform codebook with levels concentrated around the empirical bulk of \(y_t\).
The quantized value is obtained by nearest-neighbor projection: $\tilde y_t = Q(y_t) := \arg\min_{q\in\mathcal C} |y_t-q|$.

Taken together, the results in \cref{fig:y_4bit_result} suggest that Adam does not require precise pointwise fidelity in \(y_t\) in order to realize its adaptive effect. What appears to matter is not the exact value of the transformed ratio at each step,
but the coarse attenuation information that it carries. This is consistent with the distributional structure described above: most of the mass of \(y_t\) lies in a relatively compact regime, while the update depends on it only through the smooth factor \(\left(1+\frac{\beta}{1-\beta}y_t\right)^{-1/2}\). As a result, neither removing rare tail events nor coarsening the representation within the bulk of the distribution significantly perturbs the effective step size.
This robustness motivates the ratio-based reformulation developed in the next section. If Adam's essential adaptive behavior is already preserved under severe truncation and coarse quantization of the transformed ratio, then the more natural object to analyze is not the raw moment pair \((m_t,v_t)\), but the induced dynamics of the transformed ratio itself.
We emphasize that this is only a diagnostic experiment in \cref{fig:y_4bit_result}: the original Adam recursions for \(m_t\) and \(v_t\) are unchanged, and only the induced \(y_t\) is coarsened before reconstructing the denominator.

\section{Recursive Reformulation of Adam}
Unlike the diagnostic coarsening above, we now maintain \(y_t\) recursively and use it as the stored second state.
Given the stable structure of the transformed ratio, we now show that Adam can be equivalently parameterized by $(m_t, y_t)$ instead of $(m_t, v_t)$, motivating the low-precision state representation evaluated below.

\subsection{A scalar recursion for the transformed ratio}
We now derive a recursion for the transformed ratio $y_t$.
Starting from the recursion \(m_t=\beta m_{t-1}+(1-\beta)g_t\), we can rewrite the current gradient as
\begin{equation}\label{eq:g-from-x}
    g_t
    =
    \frac{m_t-\beta m_{t-1}}{1-\beta}
    =
    \frac{m_t}{1-\beta}(1-\beta x_t), \quad x_t := \frac{m_{t-1}}{m_t}.
\end{equation}
Substituting this into the second-moment recursion \(v_t=\beta v_{t-1}+(1-\beta)g_t^2\) gives
\begin{equation}\label{eq:v-recursion-x}
    v_t
    =
    \beta v_{t-1}
    +
    \frac{m_t^2}{1-\beta}(1-\beta x_t)^2.
\end{equation}
Using \(v_s=m_s^2(1+\frac{\beta}{1-\beta}y_s)\) for
\(s\in\{t,t-1\}\), together with \(m_{t-1}=x_t m_t\), 
\cref{eq:v-recursion-x} yields
\begin{equation}
\label{eq:y-recursion}
\colorbox{black!5}{$\displaystyle
\begin{aligned}
    y_t
    =
    \beta x_t^2 y_{t-1}
    +
    (1-x_t)^2.
\end{aligned}
$}
\end{equation}
Equation~\ref{eq:y-recursion} shows that, once the sign of \(m_t\) is fixed, Adam's adaptive behavior is well described by a one-dimensional transformed ratio recursion: the sign determines the update direction, while \(y_t\) controls the attenuation magnitude. Additional algebraic details are provided in Appendix~\ref{app:y-recursion-derivation}.

\begin{figure}[t]
    \centering
 \includegraphics[width=.9\linewidth]{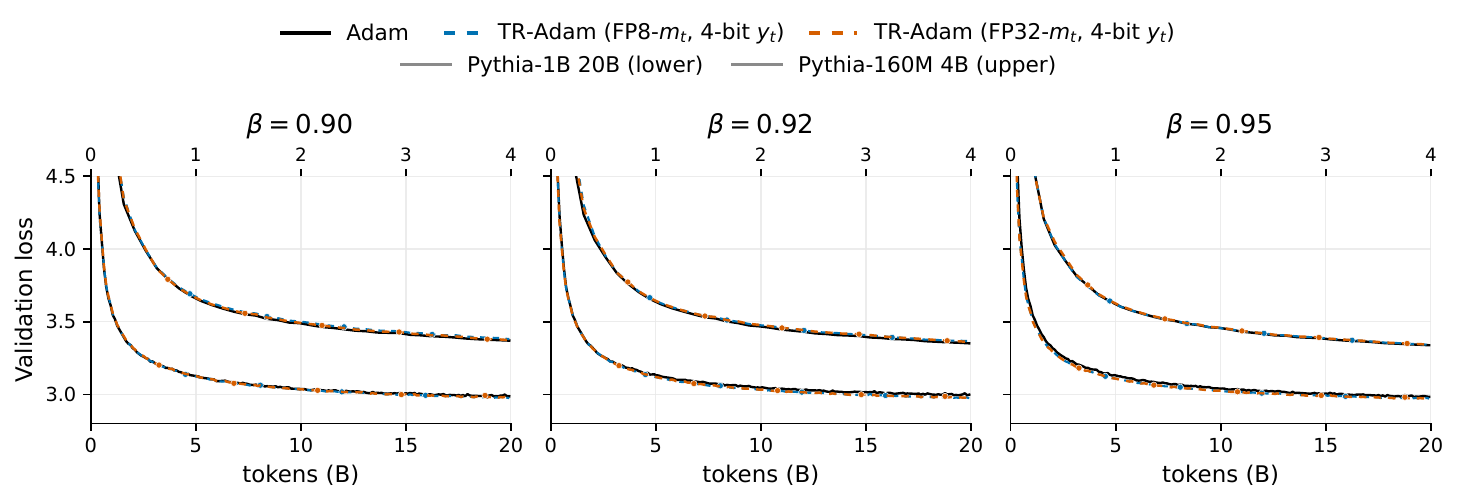}
    \caption{Pre-training comparison between full Adam and low-precision ratio representation.}
    \label{fig:ratio_quant_pretrain}
      \vspace{-1.0em}
\end{figure}

\subsection{Coarse representation through the recursion}
Equation~\ref{eq:y-recursion} suggests a natural target for coarse representation.
Instead of compressing the coupled raw states \((m_t,v_t)\) directly, we represent the transformed ratio \(y_t\), which is the scalar quantity controlling the adaptive denominator.
Concretely, after obtaining \(y_t\), we replace it by the nearest value in a fixed FP4-like codebook:
$\tilde{y}_t = Q(y_t) := \arg\min_{q\in\mathcal{C}} |y_t-q|$.
Here, \(\mathcal{C}=\{0.25,0.5,0.75,1,1.5,2,3,4,6,8,12,16,24,32,48,64\}\).
This 16-level codebook allocates more resolution to the moderate range where the transformed ratio distribution concentrates, while still retaining several larger values for the right tail.
The adaptive denominator is then reconstructed from \(\tilde y_t\) through the same transformed expression as in~\cref{eq:sign-over-ratio}, with \(y_t\) replaced by \(\tilde y_t\). 
Thus, coarse representation acts directly on the transformed-ratio attenuation rather than separately perturbing the raw moment states \(m_t\) and \(v_t\).
Notably, this representation uses no per-block scaling factors, learned quantizers, or auxiliary normalization metadata.

\subsection{Experiments on low-precision quantization of the transformed ratio}

\begin{figure}[t]
   \vspace{-1.5em}
    \centering
    \includegraphics[width=.94\linewidth]{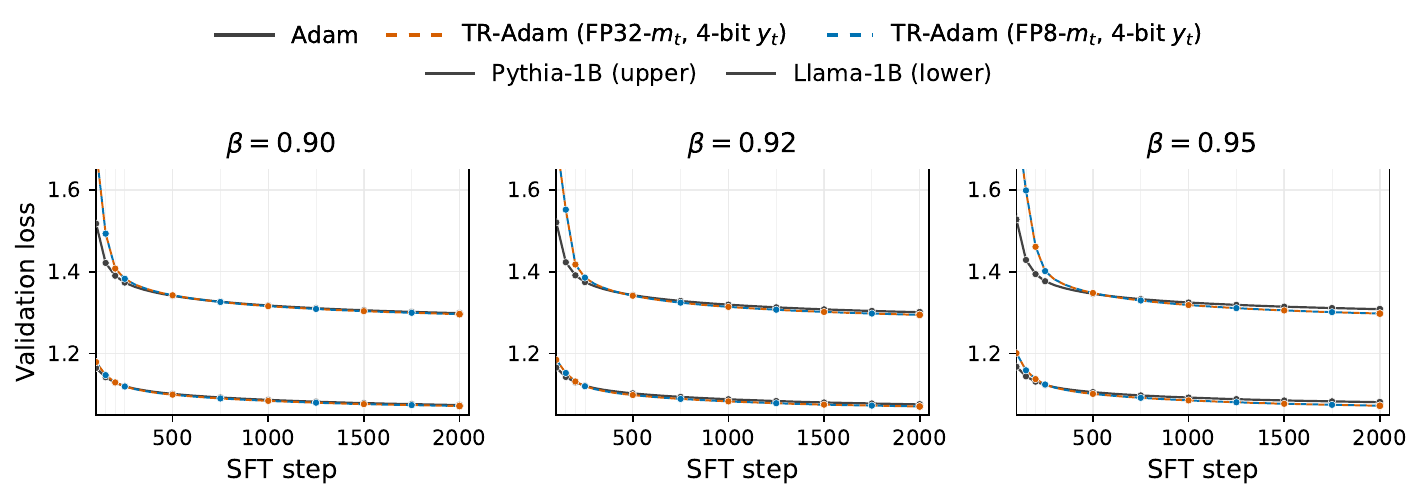}
    \caption{SFT comparison between full Adam and low-precision transformed-ratio representation.}
    \label{fig:ratio_quant_sft}
\end{figure}

\begin{figure}[h]
  \vspace{-1.0em}
    \centering
    \includegraphics[width=.94\linewidth]{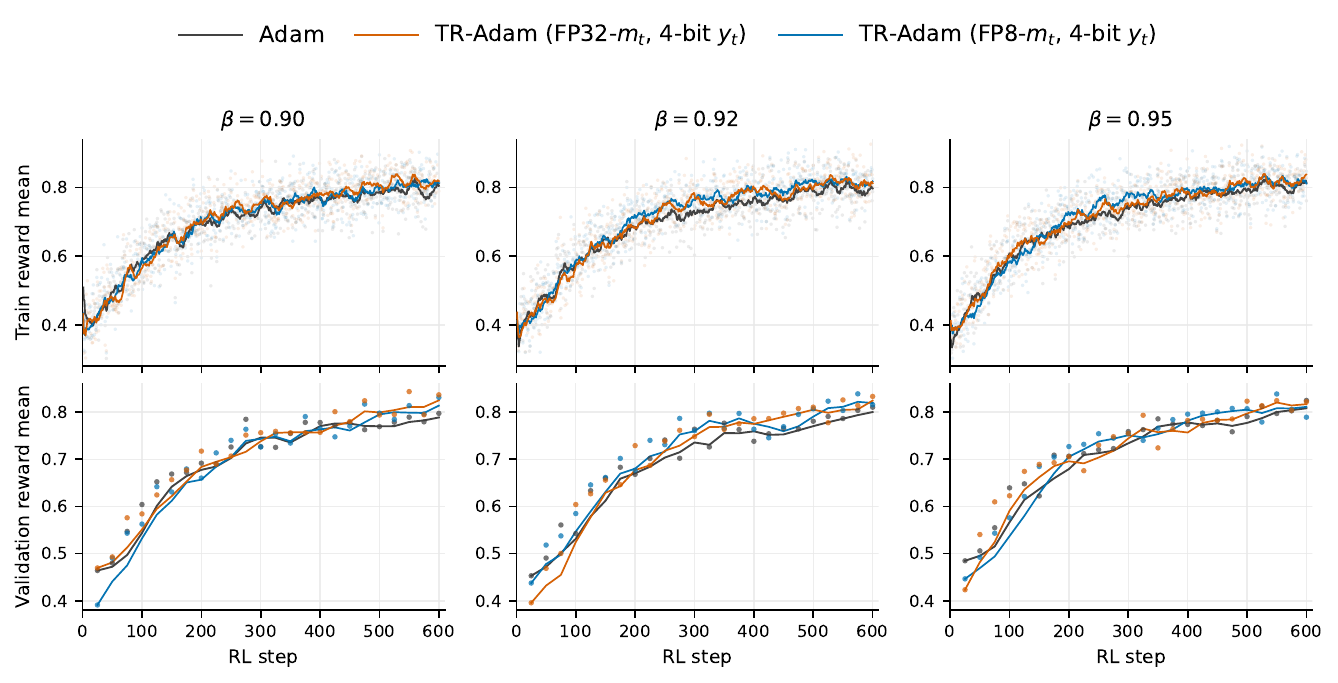}
    \caption{RLHF comparison between full Adam and low-precision transformed-ratio representation.}
    \label{fig:ratio_quant_rlhf}
    \vspace{-1.5em}
\end{figure}

We next examine whether the transformed ratio representation remains effective across the full language-model training pipeline. 
We compare tied-\(\beta\) Adam with its low-precision transformed ratio variant in three representative stages: pre-training, supervised fine-tuning (SFT), and reinforcement-learning-based alignment (RLHF). 
The training recipe is held fixed while we vary the representation of the ratio and first-moment states. 
In addition to storing \(y_t\) with a 4-bit codebook, we also evaluate variants in which the first-moment state \(m_t\) is stored in FP8~\citep{peng2023fp8lm,chitsaz2024exploring,fishman2025scalingfp8}.
Low-precision states are emulated in FP32 tensors: these experiments assess quantization effects, not packed-state memory or runtime savings. This is a useful stress test for the transformed-ratio formulation: although the recursion depends on ratios involving \(m_t\), the combination of FP8 first-moment storage and 4-bit transformed-ratio storage still preserves strong performance in our experiments.
\begin{figure}[t]
    \centering
    \includegraphics[width=0.70\linewidth]{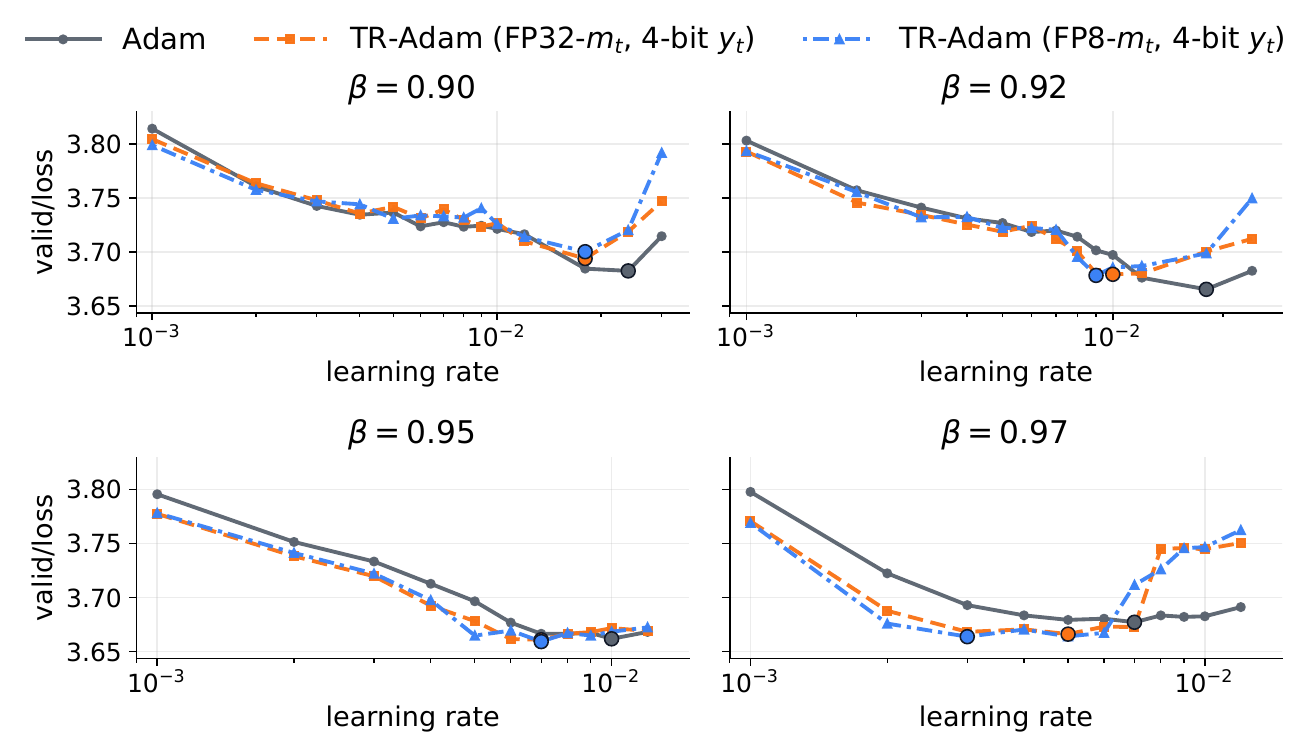}
    \caption{
    Learning-rate sweeps on Llama-20M pre-training comparing tied-\(\beta\) Adam with transformed-ratio Adam using FP32 or FP8 first-moment storage and 4-bit \(y_t\).
    }
    \label{fig:llama20m_lr_sweep_fp32_fp8}
\end{figure}
\Cref{fig:ratio_quant_pretrain,fig:ratio_quant_sft,fig:ratio_quant_rlhf} show that the low-precision transformed-ratio variant closely tracks full Adam across all three stages. In pre-training, the two validation-loss curves nearly overlap throughout optimization and reach essentially the same final loss, suggesting that the transformed ratio preserves the denominator-side information needed for long-horizon large-scale training. In SFT and RLHF, the low-precision variant may initially lag the full-Adam baseline, but it improves steadily and often matches or slightly exceeds Adam later in training. Thus, the effect of coarse transformed-ratio representation is not merely harmless compression: in post-training regimes, it can behave like a mild smoothing or regularization of the adaptive denominator dynamics.

We further examine whether the low-precision transformed-ratio representation preserves Adam's learning-rate sensitivity. 
\Cref{fig:llama20m_lr_sweep_fp32_fp8} shows learning-rate sweeps on a Llama-20M pre-training setup under four tied-\(\beta\) values.
Across \(\beta=0.90,0.92,0.95,0.97\), both transformed-ratio variants closely track the Adam sweep profile: their favorable learning-rate regions remain aligned with Adam, and performance degradation at overly large learning rates occurs at similar scales. 
This indicates that coarse transformed-ratio storage does not simply succeed after careful retuning at a single operating point; rather, it largely preserves Adam's effective learning-rate structure.

Overall, the results show that the transformed ratio \(y_t\) is a robust target for coarse representation: in pre-training it recovers Adam-like behavior almost exactly, while in SFT and RLHF it remains competitive and can sometimes improve late-stage performance. 
This stage-consistent behavior suggests that the essential adaptive dynamics of tied-\(\beta\) Adam are captured by a low-complexity transformed-ratio structure rather than by high-precision values of the raw second-moment state. 
More experimental results are provided in Appendix~\ref{more_exp}.

\WFclear
\section{The Signum-like Limit}

\subsection{A constant-state approximation and the Signum-like limit}

The transformed-ratio representation provides a simple interpretation of Adam as a sign-momentum method with stochastic attenuation.
From~\cref{eq:sign-over-ratio}, once the sign of the momentum is fixed, Adam's adaptive behavior is controlled by the scalar attenuation factor \((1+(\beta/(1-\beta))y_t)^{-1/2}\). 
This motivates a constant-state approximation in which the transformed ratio \(y_t\) is replaced by a representative constant value \(y_{\mathrm{const}}\ge 0\). 
The resulting Signum-type update and the corresponding learning-rate matching rule are
\begin{equation}
    \Delta\theta_t
    \approx
    -\eta_{\mathrm{adam}}
    \left(1+\frac{\beta}{1-\beta}y_{\mathrm{const}}\right)^{-\frac{1}{2}}
    \operatorname{sign}(m_t),
    \quad
    \eta_{\mathrm{sign}}
    \approx
    \eta_{\mathrm{adam}}
    \left(1+\frac{\beta}{1-\beta}y_{\mathrm{const}}\right)^{-\frac{1}{2}}.
    \label{eq:sign-lr-match-general}
\end{equation}
Here, the first expression is the constant-\(y_t\) Signum-type update, while the second gives the learning-rate scale needed to write it as
\(\Delta\theta_t^{\mathrm{sign}}=-\eta_{\mathrm{sign}}\operatorname{sign}(m_t)\).
This relation should be viewed as a matching rule: the full Adam update still retains the coordinate-wise fluctuations and temporal variation of \(y_t\), whereas the constant-state approximation preserves only its dominant effective scale.

To choose \(y_{\mathrm{const}}\), we match the expected attenuation, or equivalently the expected effective step size, under the transformed-ratio distribution:
\begin{align}
    \mathbb{E}\!\bigl[
    (1+\tfrac{\beta}{1-\beta}y_t)^{-\frac{1}{2}}
    \bigr]
    =
    (1+\tfrac{\beta}{1-\beta}y_{\mathrm{const}})^{-\frac{1}{2}}.
    \label{eq:yconst-step-match}
\end{align}
Solving for the attenuation-matched constant gives
\begin{align}
    y_{\mathrm{const}}
    =
    \frac{1-\beta}{\beta}
    \Big(
    \mathbb{E}\!\big[
    (1+\tfrac{\beta}{1-\beta}y_t)^{-\frac{1}{2}}
    \big]^{-2}
    -1
    \Big).
    \label{eq:yconst-solved}
\end{align}

This calibration is useful because it connects the constant-state approximation directly to Adam's native learning-rate scale, rather than treating Signum as requiring an unrelated learning-rate search.

To make the scale of \(y_{\mathrm{const}}\) more concrete, we substitute the shifted inverse-square reference \(Y_c=2/(Z+c)^2\) from Proposition~\ref{thm:y-structural-decomposition} into this expression.
This yields
\begin{align}
    y_{\mathrm{const}}^{\mathrm{eff}}(\beta,c)
    =
    \frac{1-\beta}{\beta}
    \left(
    \mathbb{E}_{Z\sim\mathcal{N}(0,1)}
    \bigl[
    (1+\tfrac{2\beta}{(1-\beta)(Z+c)^2})^{-\frac{1}{2}}
    \bigr]^{-2}
    -1
    \right).
    \label{eq:yconst-effective-shifted-reference}
\end{align}

Numerically, this value is stable for the small shifts considered in the distributional model. 
As shown in \cref{tab:yconst-effective}, the attenuation-matched constant remains slightly above \(3\) across \(\beta\in[0.90,0.97]\) and \(c\in\{0,0.1,0.2\}\).
Thus, under the same shifted-reference law used to describe the empirical transformed-ratio distribution, a moderate constant transformed-ratio value should already capture Adam's typical attenuation scale. 

A practical advantage of this view is that it expresses a Signum-like optimizer in Adam's native learning-rate parameterization. 
In modern language-model training, Adam learning-rate heuristics are well established, whereas Signum often requires a separate learning-rate search. 
Replacing \(y_t\) with a moderate constant therefore yields a simplified sign-style update that remains calibrated to the Adam learning-rate scale, making it possible to reuse Adam-style tuning rather than introducing a separate optimizer-specific search.
We therefore test several representative values of \(y_{\mathrm{const}}\) and compare them against Adam and Signum in learning-rate sweeps across different values of \(\beta\).

\subsection{Learning-rate sweeps across \texorpdfstring{\(\beta\)}{beta}: empirical observations}

\begin{wrapfigure}{r}{0.60\textwidth}
    \vspace{-1.8em}
    \centering
    \includegraphics[width=\linewidth]{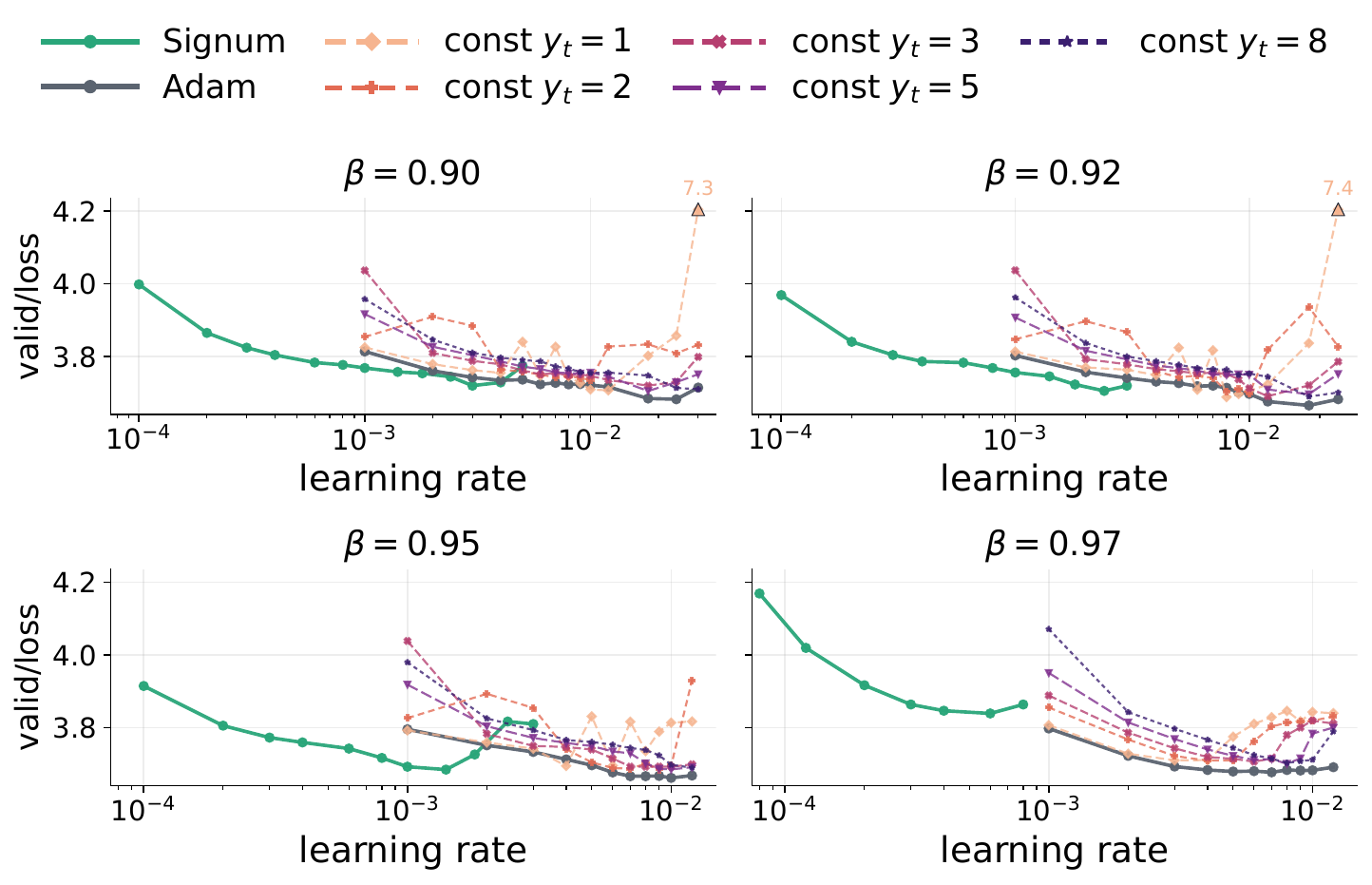}
    \caption{
Learning-rate sweeps: 
Signum shifts to smaller learning rates as \(\beta\) increases, while constant-state proxies \(y_{\mathrm{const}}\in\{3,5\}\) best preserve Adam's learning-rate scale.
}\label{fig:signum_adam_lr_sweeps}
    \vspace{-1.2em}
\end{wrapfigure}

Using the Llama-style 20M model, we now test the constant-state proxy by comparing learning-rate sweeps of Adam, Signum, and several constant-\(y_t\) variants across a range of \(\beta\); see~\cref{fig:signum_adam_lr_sweeps}. 
Two empirical trends consistently appear.

First, Signum's preferred learning-rate region shifts systematically to smaller values as \(\beta\) increases. 
This behavior is consistent with the learning-rate scale predicted by~\cref{eq:sign-lr-match-general}. 
If Adam's preferred learning-rate scale is viewed as relatively stable across \(\beta\), then a pure Signum update must compensate for the transformed-ratio attenuation that Adam applies. 
Since the prefactor \(\beta/(1-\beta)\) grows rapidly with \(\beta\), the effective attenuation in~\cref{eq:sign-lr-match-general} becomes stronger at larger \(\beta\). 
Thus, matching Adam's effective step size requires a smaller Signum learning rate. 
This matches the trend observed empirically in~\cref{fig:signum_adam_lr_sweeps}: as \(\beta\) increases, the favorable learning-rate region for Signum moves markedly left relative to Adam.
\begin{table}[tb]
 \vspace{-1em}
\centering
\footnotesize
\setlength{\tabcolsep}{2.8pt}
\renewcommand{\arraystretch}{0.86}
\caption{\(y_{\mathrm{const}}^{\mathrm{eff}}\) under \(Y_c=2/(Z+c)^2\).}
\label{tab:yconst-effective}
\begin{tabular}{c@{\quad}cccc}
\toprule
\(c\) & \(\beta=0.90\) & \(\beta=0.92\) & \(\beta=0.95\) & \(\beta=0.97\) \\
\midrule
0   & 3.36 & 3.31 & 3.25 & 3.21 \\
0.1 & 3.33 & 3.28 & 3.22 & 3.17 \\
0.2 & 3.23 & 3.19 & 3.13 & 3.08 \\
\bottomrule
\end{tabular}
\vspace{-1.0em}
\end{table}
Second, the constant-\(y_t\) variants preserve Adam's learning-rate scale much better than pure Signum. 
Among the tested constants, \(y_{\mathrm{const}}\in\{3,5\}\) gives the closest overall match to Adam's preferred learning-rate region.
This is broadly consistent with the step-size-matching calculation in~\cref{eq:yconst-effective-shifted-reference} and the numerical values reported above, which place the effective constant slightly above \(3\) under the shifted-reference model. 
Although the true transformed ratio \(y_t\) is stochastic and heavy-tailed, these results suggest that its dominant effect on the optimizer's learning-rate scale can be captured by a single moderate constant.

This observation is practically useful. 
Directly using Signum requires a separate learning-rate search, and the appropriate learning rate changes noticeably with \(\beta\). 
By contrast, the constant-\(y_t\) proxy remains in Adam's native learning-rate parameterization: replacing \(y_t\) by a moderate constant yields a Signum-like update whose effective step size is already calibrated to Adam. 
Taken together, the sweeps show that tied-\(\beta\) Adam behaves like a sign-based method modulated by the transformed-ratio attenuation, and that a constant approximation can preserve much of Adam's learning-rate scale.

\section{Conclusion and Limitations}
In this paper, we studied Adam in the tied-\(\beta\) regime, \(\beta_1=\beta_2\), through a transformed ratio \(y_t\). 
By factoring out the explicit \(\beta\)-dependent scale, this representation exposes a comparatively stable ratio structure in Adam's adaptive denominator that is obscured in the raw moment states \(m_t\) and \(v_t\). 
Empirically, the distribution of \(y_t\) remains consistent across training settings: most coordinates concentrate in a moderate range, while a persistent heavy tail captures rare large ratio-state values.
By isolating an approximately stable state inside Adam's adaptive denominator, the ratio view yields two practical consequences.
First, \(y_t\) provides a compact replacement for the raw second-moment state \(v_t\): across the language-model training regimes studied here, a fixed 4-bit codebook for \(y_t\) preserves performance close to that of full-precision Adam.
Second, the same representation clarifies Adam's connection to Signum: tied-\(\beta\) Adam can be viewed as a sign-momentum update attenuated by \(y_t\), and replacing \(y_t\) by a constant yields a Signum-type limit with an explicit learning-rate scaling. 
Overall, these results suggest that tied-$\beta$ Adam's adaptive behavior can be captured to a surprising extent through coarse but stable ratio information.
Our main optimizer-performance evaluation covers Transformer language models up to 1B parameters; Appendix~\ref{app:architecture-scale} adds short-budget MLP and ResNet experiments and an early-stage 8B ratio-distribution check. Full-training optimizer comparisons beyond 1B and broader architecture coverage remain future work. The additional seed, decay-rate, and numerical-stability studies in Appendix~\ref{more_exp} further delimit the scope of the empirical claims.

\begin{ack}
This work was supported in part by the Engineering and Physical Sciences Research Council (EPSRC) through the AI Hub in Generative Models [grant number EP/Y028805/1].
Samuel Kaski is supported by UKRI Turing AI World-Leading Researcher Fellowship, EP/W002973/1 and EU, ERC ODD-ML 101201120.
\end{ack}

\clearpage

\bibliographystyle{plainnat}
\bibliography{references}

\clearpage

\appendix
\crefname{section}{Appendix}{Appendices}
\Crefname{section}{Appendix}{Appendices}
\listofappendix

\input{appendix}

\clearpage
\input{checklist}

\end{document}

%% file: appendix.tex
\appendixsection{Distributional properties of the transformed ratio state}
\label{app:y-distribution}

Appendix~\ref{app:y-distribution} records the derivation behind~\cref{thm:y-structural-decomposition}. The main text states the local
Gaussian model and the formal result; here we keep only the short intuition
needed to read the proof. In the tied-beta regime, the transformed ratio state is
\[
    y_t
    =
    \frac{1-\beta}{\beta}\frac{v_t-m_t^2}{m_t^2}.
\]
Under a local Gaussian window, the EMA denominator has the form
\(m_t=\sigma_m(Z+a)\), so the small-denominator event \(Z+a\approx0\) naturally
produces the shifted inverse-square reference \(2/(Z+a)^2\). The numerator does
not disappear; the proof below shows that it enters as the explicit relative
correction \(\eta_t\) in~\cref{eq:yt-relative-reference}.

When the local shift varies across coordinates or nearby windows, let
\(\mathcal I\) index the observed coordinate-window instances and let each
instance contribute its shift \(a_i\). Flattening these shifts gives the
empirical pooled-shift distribution
\[
    \widehat P_A
    =
    \frac{1}{|\mathcal I|}\sum_{i\in\mathcal I}\delta_{a_i},
    \qquad
    A\sim\widehat P_A .
\]
Thus \(A\sim\widehat P_A\) is the shift obtained by selecting one observed
instance uniformly. More generally, for a pooled-shift distribution \(P\),
drawing \(A\sim P\) gives the mixed reference \(Y_A=2/(Z+A)^2\). The fixed
reference \(Y_c=2/(Z+c)^2\), with \(c^2:=\mathbb E_P[A^2]\), keeps the leading
shift effect; the shifted-reference calculation below proves the corresponding
second-order survival matching.

\paragraph{Proof of~\cref{thm:y-structural-decomposition} and shifted-reference calculation.}
We justify~\cref{thm:y-structural-decomposition} under the local Gaussian
window used in the main text. The calculation is not a global model of training;
it isolates the inverse-square denominator that controls the ratio state.

\emph{Exact factorization.}
Let \(L=\sum_{j\ge0}w_j\xi_j\), \(Q=\sum_{j\ge0}w_j\xi_j^2\), and
\(s_\beta^2=\sum_{j\ge0}w_j^2=(1-\beta)/(1+\beta)\). Then
\begin{align*}
    m_t &= \mu+\sigma L=\sigma_m(Z+a),\\
    U_t &= v_t-m_t^2=\sigma^2(Q-L^2),\qquad Z=L/s_\beta .
\end{align*}
Since \(\mathbb E[Q]=1\) and \(\mathbb E[L^2]=s_\beta^2\),
\begin{align*}
    \mathbb E[U_t]
    &= \sigma^2(1-s_\beta^2)=\frac{2\beta}{1+\beta}\sigma^2,\\
    \frac{1-\beta}{\beta}\frac{\mathbb E[U_t]}{\sigma_m^2}
    &= 2 .
\end{align*}
Now decompose the numerator around the denominator coordinate,
\[
    U_t=\mathbb E[U_t]
    +\bigl(\mathbb E[U_t\mid Z]-\mathbb E[U_t]\bigr)
    +\bigl(U_t-\mathbb E[U_t\mid Z]\bigr).
\]
For Gaussian variables \(\xi_j\), projecting onto \(Z\) gives
\begin{align*}
    \mathbb E[U_t\mid Z]-\mathbb E[U_t]
    &= \sigma^2 d_\beta (Z^2-1),\\
    d_\beta
    &= \frac{\sum_{j\ge0}w_j^3}{s_\beta^2}-s_\beta^2
     = \frac{\beta(1-\beta)}{(1+\beta)(1+\beta+\beta^2)} .
\end{align*}
Substituting into \(y_t=\frac{1-\beta}{\beta}U_t/m_t^2\) yields
\begin{align*}
    y_t
    &= \frac{2+\eta_t}{(Z+a)^2}
     = Y_a\left(1+\frac{\eta_t}{2}\right),\\
    \eta_t
    &= \eta_t^{\mathrm{coup}}+\eta_t^{\mathrm{fluc}},\\
    \eta_t^{\mathrm{coup}}
    &= \gamma_\beta(Z^2-1),
    \qquad
    \gamma_\beta=\frac{1-\beta}{1+\beta+\beta^2},\\
    \eta_t^{\mathrm{fluc}}
    &= \frac{1-\beta}{\beta}
       \frac{U_t-\mathbb E[U_t\mid Z]}{\sigma_m^2},
\end{align*}
where \(Y_a=2/(Z+a)^2\). 
This proves the factorization in~\cref{eq:yt-relative-reference}.

\emph{Auxiliary correction estimates.}
The fluctuation term is conditionally centered by construction:
\[
    \mathbb E[\eta_t^{\mathrm{fluc}}\mid Z]=0 .
\]
For its variance, write \(B=\operatorname{diag}(w)-ww^\top\), so that
\(U_t=\sigma^2\xi^\top B\xi\). The Gaussian quadratic-form identity gives
\[
    \operatorname{Var}(U_t)
    =
    2\sigma^4\operatorname{tr}(B^2)
    =
    2\sigma^4\left(s_\beta^2-2\sum_{j\ge0}w_j^3+s_\beta^4\right).
\]
Since
\(\mathbb E[U_t\mid Z]-\mathbb E[U_t]=\sigma^2d_\beta(Z^2-1)\),
\(\operatorname{Var}(\mathbb E[U_t\mid Z])=2\sigma^4d_\beta^2\), we have
\[
    \operatorname{Var}(\eta_t^{\mathrm{fluc}})
    =
    \frac{2(1+\beta)^2}{\beta^2}
    \left(
        s_\beta^2-2\sum_{j\ge0}w_j^3+s_\beta^4-d_\beta^2
    \right)
    =
    \frac{2(1+\beta)}{\beta^2}(1-\beta)+O((1-\beta)^2).
\]
Also, since \(y_t/Y_a-1=\eta_t/2\), for any \(0<u<2\delta\),
\[
    \left\{\left|\frac{y_t}{Y_a}-1\right|>\delta\right\}
    \subseteq
    \{|\eta_t^{\mathrm{coup}}|>u\}
    \cup
    \{|\eta_t^{\mathrm{fluc}}|>2\delta-u\}.
\]
Taking probabilities of both sides and applying Chebyshev's inequality gives the corresponding
relative-error control.

\emph{Tail order.}
Let \(S_t=2+\eta_t\ge0\). Then
\[
    \Pr(y_t>k)
    =
    \int_{\mathbb R}
    \phi(z)\,
    \Pr\!\left(S_t>k(z+a)^2\mid Z=z\right)\,dz .
\]
With \(x=\sqrt{k}(z+a)\), this becomes
\[
    \frac{1}{\sqrt{k}}
    \int_{\mathbb R}
    \phi\!\left(-a+\frac{x}{\sqrt{k}}\right)
    \Pr\!\left(S_t>x^2\mid Z=-a+\frac{x}{\sqrt{k}}\right)\,dx .
\]
We justify domination using the Gaussian quadratic-form structure. Set
\(\alpha=w/s_\beta\), \(C=(1-\beta)/(\beta s_\beta^2)\), and decompose
\(\xi=\alpha Z+\xi_\perp\), with \(\xi_\perp\) independent of \(Z\).
Since \(B=\operatorname{diag}(w)-ww^\top\) is positive semidefinite and trace class,
\(V=\sqrt C B^{1/2}\xi_\perp\) has finite second moment, and
\(d=\sqrt C B^{1/2}\alpha\) has finite norm. Conditionally on \(Z=z\),
\(\sqrt{S_t}\) has the representation \(\|V+dz\|\).
Consequently \(h(z)=\mathbb E\|V+dz\|\) is finite and Lipschitz continuous.
Let \(D=\|V\|+|a|\|d\|\). For \(\sqrt{k}\ge2\|d\|\),
\(\|V+d(-a+x/\sqrt{k})\|>|x|\) implies \(D>|x|/2\).
The integrand above is therefore bounded by the integrable function
\[
 (2\pi)^{-1/2}\min\{1,4\mathbb E[D^2]/x^2\},
\]
with its value at zero defined by continuity. The conditional tail probabilities converge for almost every \(x\), since \(\|V+d(-a+x/\sqrt{k})\|\to\|V-da\|\) almost surely. Dominated convergence now gives
\[
    \Pr(y_t>k)
    =
    \frac{\phi(a)}{\sqrt{k}}
    \int_{\mathbb R}\Pr(S_t>x^2\mid Z=-a)\,dx
    +o(k^{-1/2})
    =
    \frac{2\phi(a)}{\sqrt{k}}h(-a)+o(k^{-1/2}).
\]

\emph{Shifted-reference matching.}
Let \(P\) be a pooled-shift distribution with finite fourth moment, and draw
\(A\sim P\). If \(P\) is empirical, then
\(P=|\mathcal I|^{-1}\sum_{i\in\mathcal I}\delta_{a_i}\), and
\[
    \mathbb E_P[A^2]
    =
    \frac{1}{|\mathcal I|}\sum_{i\in\mathcal I}a_i^2 .
\]
Assume \(Z\perp A\) and \(Z\sim\mathcal N(0,1)\). Fix \(k>0\). Writing
\(q=\sqrt{2/k}\), the mixed denominator reference \(Y_A=2/(Z+A)^2\) satisfies
\[
    \Pr(Y_A>k\mid A)
    =
    \Phi(q-A)+\Phi(q+A)-1 .
\]
A Taylor expansion in the shift gives
\[
    \Phi(q-A)+\Phi(q+A)-1
    =
    2\Phi(q)-1-q\phi(q)A^2+O(A^4).
\]
Therefore
\begin{align*}
    \Pr(Y_A>k)
    &=
    2\Phi(q)-1-q\phi(q)\mathbb E_P[A^2]
    +O(\mathbb E_P[A^4]),\\
    \Pr(Y_c>k)
    &=
    2\Phi(q)-1-q\phi(q)c^2+O(c^4),
    \qquad c^2:=\mathbb E_P[A^2].
\end{align*}
For a family \(P_\varepsilon\) of pooled-shift distributions, let
\(A_\varepsilon\sim P_\varepsilon\) and
\(c_\varepsilon^2:=\mathbb E_{P_\varepsilon}[A_\varepsilon^2]\). If
\(\mathbb E_{P_\varepsilon}[A_\varepsilon^4]=O(\varepsilon^4)\), Jensen's
inequality gives
\(c_\varepsilon^4=(\mathbb E_{P_\varepsilon}[A_\varepsilon^2])^2\le
\mathbb E_{P_\varepsilon}[A_\varepsilon^4]=O(\varepsilon^4)\). Hence the two
survival functions differ only at fourth order:
\[
    \Pr(Y_{A_\varepsilon}>k)-\Pr(Y_{c_\varepsilon}>k)=O(\varepsilon^4).
\]
This proves the second-order matching for the reference \(Y_c\) in~\cref{eq:yc-reference}.

\appendixsection{Additional Motivation and Related Work for the Tied-\texorpdfstring{$\beta$}{beta} Regime}
\label{app:balanced_regime_motivation}

This appendix provides additional motivation for studying the tied-\(\beta\) regime, \(\beta_1=\beta_2\). Our goal is not to claim that this choice is universally optimal, but rather to explain why it is a structurally distinguished regime for analyzing Adam. The classical Adam default \((\beta_1,\beta_2)=(0.9,0.999)\) uses very different memories for the first and second moving averages. However, recent analyses and large-scale training practice have increasingly considered settings where these two memories are closer, including tied or nearly tied choices. This motivates asking what becomes simpler or more stable when the two exponential memories are matched.

Several recent works point in this direction from complementary perspectives. 
\citet{orvieto2025adamsecret} show that tied-\(\beta\) Adam preserves much of Adam's empirical behavior in language-model training while admitting a cleaner interpretation in terms of smoothed sign momentum and a noise-to-signal correction. 
\citet{fernandezhernandez2026adamscale} identify a structural principle behind this regime, showing that Adam has first-order gradient-scale invariance precisely when \(\beta_1=\beta_2\). 
A separate line of work studies how mini-batch noise interacts with Adam's memory parameters, reporting that moving \(\beta_1\) closer to \(\beta_2\) can improve validation behavior in some larger-batch, multi-epoch settings~\citep{cattaneo2026minibatchnoiseadam}.
A concise informal overview of several related perspectives on the tied-\(\beta\) regime is also provided by \citet{kexuefm2026betaeq}. 
These results suggest that tied or near-tied memories are not merely algebraically convenient, but can reveal meaningful structure in Adam's adaptive dynamics.

Our use of the tied-\(\beta\) regime is more specific. 
It is exactly the setting in which Adam's adaptive denominator admits a clean sign-over-ratio reformulation with a nonnegative variance-like residual. 
Ignoring bias correction and \(\epsilon\) for simplicity, Adam evolves coordinate-wise as
\begin{align}
    m_t &= \beta_1 m_{t-1} + (1-\beta_1) g_t, \\
    v_t &= \beta_2 v_{t-1} + (1-\beta_2) g_t^2,
\end{align}
and the adaptive factor is \(m_t/\sqrt{v_t}\). 
When \(\beta_1=\beta_2=\beta\), a direct expansion gives
\begin{align}
    v_t - m_t^2
    =
    \beta\bigl(v_{t-1}-m_{t-1}^2\bigr)
    + \beta(1-\beta)(g_t-m_{t-1})^2 .
    \label{eq:tied-beta-nonnegative-recursion}
\end{align}
Therefore, under the standard zero initialization,
\begin{align}
    v_t - m_t^2 \ge 0
    \qquad \text{for all } t .
\end{align}
This makes \(v_t-m_t^2\) a genuine variance-like residual. 
Whenever \(m_t\neq 0\), we can define
\begin{align}
    r_t := \frac{v_t-m_t^2}{m_t^2} \ge 0,
    \qquad
    v_t = m_t^2(1+r_t),
\end{align}
which yields
\begin{align}
    \frac{m_t}{\sqrt{v_t}}
    =
    \frac{\operatorname{sign}(m_t)}{\sqrt{1+r_t}} .
\end{align}
Thus, in the tied-\(\beta\) regime, Adam's adaptive factor is a sign-momentum direction multiplied by an attenuation factor. 
Since \(r_t\ge 0\), this attenuation is at most one; the update cannot be amplified beyond the corresponding sign-momentum magnitude by a negative residual.

This nonnegativity property is in fact special to the tied-\(\beta\) regime. 
For \(0<\beta_1,\beta_2<1\), fix a time \(t\) and write the zero-initialized moment estimates as
\begin{align}
    m_t
    &=
    \sum_{i=0}^{t-1} a_i g_{t-i},
    \qquad
    a_i := (1-\beta_1)\beta_1^i, \\
    v_t
    &=
    \sum_{i=0}^{t-1} b_i g_{t-i}^2,
    \qquad
    b_i := (1-\beta_2)\beta_2^i .
\end{align}
By weighted Cauchy--Schwarz,
\begin{align}
    m_t^2
    =
    \left(\sum_{i=0}^{t-1} a_i g_{t-i}\right)^2
    \le
    \left(\sum_{i=0}^{t-1}\frac{a_i^2}{b_i}\right)
    \left(\sum_{i=0}^{t-1}b_i g_{t-i}^2\right)
    =
    S_t v_t,
    \label{eq:weighted-cauchy-mv}
\end{align}
where
\begin{align}
    S_t
    :=
    \sum_{i=0}^{t-1}\frac{a_i^2}{b_i}
    =
    \frac{(1-\beta_1)^2}{1-\beta_2}
    \sum_{i=0}^{t-1}
    \left(\frac{\beta_1^2}{\beta_2}\right)^i .
    \label{eq:st-condition}
\end{align}
Moreover, the bound is tight: equality in~\cref{eq:weighted-cauchy-mv} is attained by choosing \(g_{t-i}\propto a_i/b_i\). 
Consequently, \(v_t\ge m_t^2\) for all gradient sequences at time \(t\) holds if and only if \(S_t\le 1\).

When \(\beta_1=\beta_2=\beta\), \cref{eq:st-condition} gives
\begin{align}
    S_t
    =
    (1-\beta)\sum_{i=0}^{t-1}\beta^i
    =
    1-\beta^t
    \le 1,
\end{align}
recovering the nonnegativity above. 
Conversely, suppose \(\beta_1\ne\beta_2\). 
If \(\beta_1^2\ge \beta_2\), then the sum in~\cref{eq:st-condition} grows without bound as \(t\) increases, so \(S_t>1\) for some finite \(t\). 
If \(\beta_1^2<\beta_2\), then
\begin{align}
    \lim_{t\to\infty} S_t
    =
    \frac{\beta_2(1-\beta_1)^2}
    {(1-\beta_2)(\beta_2-\beta_1^2)} .
\end{align}
A direct comparison gives
\begin{align}
    \frac{\beta_2(1-\beta_1)^2}
    {(1-\beta_2)(\beta_2-\beta_1^2)}
    > 1
    \qquad
    \Longleftrightarrow
    \qquad
    (\beta_2-\beta_1)^2>0,
\end{align}
which holds whenever \(\beta_1\ne\beta_2\). 
Thus \(S_t>1\) for some finite \(t\). 
By the tightness of~\cref{eq:weighted-cauchy-mv}, there exists a gradient sequence for which \(m_t^2/v_t=S_t>1\), and therefore
\begin{align}
    v_t-m_t^2<0 .
\end{align}
Hence the tied-\(\beta\) regime is not only sufficient but necessary for the residual \(v_t-m_t^2\) to remain nonnegative for all gradient sequences.

A simple spike calculation illustrates this failure mode for the classical mismatched setting. 
For fixed previous state \((m_{t-1},v_{t-1})\),
\begin{align}
    1+r_t
    =
    \frac{v_t}{m_t^2}
    =
    \frac{\beta_2 v_{t-1} + (1-\beta_2) g_t^2}
    {\bigl(\beta_1 m_{t-1} + (1-\beta_1) g_t\bigr)^2}.
\end{align}
As \(|g_t|\to\infty\), the lower-order terms vanish and
\begin{align}
    1+r_t
    \longrightarrow
    \frac{1-\beta_2}{(1-\beta_1)^2}.
\end{align}
For the classical Adam default \((\beta_1,\beta_2)=(0.9,0.999)\), this limiting value is
\begin{align}
    \frac{1-\beta_2}{(1-\beta_1)^2}
    =
    \frac{0.001}{0.1^2}
    =
    0.1.
\end{align}
So in the large-spike regime \(1+r_t\approx 0.1\).
Thus, away from the tied-\(\beta\) regime, the sign-over-ratio residual is not protected by the same nonnegativity mechanism and can become substantially negative.

This observation also connects to work on nonstationary training dynamics, especially in reinforcement learning. In such settings, gradient statistics can shift rapidly, and mismatched first- and second-moment memories may cause the adaptive denominator to lag behind the signal term. This can temporarily weaken the attenuation and produce unexpectedly large updates. Related reinforcement-learning studies have connected Adam's timescale choices to policy collapse or instability under nonstationary data streams, and report that matching the first- and second-moment decay rates, or bringing them closer together, can mitigate some of these failures~\citep{dohare2023policycollapse,moalla2024norepresentation}. From another angle, \citet{ellis2024adamrel} show that nonstationary changes in gradient magnitude can cause Adam to produce overly large updates, motivating a relative-timestep correction. These works do not establish tied \(\beta\) values as a universally optimal choice, but they support the broader view that the relative timescales of Adam's two moving averages are central to stability under changing gradient statistics.

Taken together, these perspectives motivate the tied-\(\beta\) regime as a useful analytical setting. It is the regime in which the adaptive denominator admits a clean decomposition into a squared mean and a nonnegative variance-like residual; it aligns with recent structural analyses of Adam's scale behavior; and it provides a natural starting point for our ratio-state formulation.

\appendixsection{Details for the transformed-state recursion and implementation}
\label{app:y-recursion-derivation}

This appendix collects additional details for the transformed-state recursion used in the main text. We first give a step-by-step derivation of the scalar recursion for \(y_t\), then describe its initialization and the practical handling of numerically unstable regions, and finally summarize the ablations used to evaluate the ratio-recursive formulation.

\subsection{Step-by-step derivation of the transformed-state recursion}

Recall that in the tied-beta regime \(\beta_1=\beta_2=\beta\), we define
\begin{align}
    r_t := \frac{v_t-m_t^2}{m_t^2},
    \qquad
    y_t := \frac{1-\beta}{\beta} r_t
    = \frac{1-\beta}{\beta}\frac{v_t-m_t^2}{m_t^2},
\end{align}
whenever \(m_t\neq 0\). Equivalently,
\begin{align}
    v_t = m_t^2\left(1+\frac{\beta}{1-\beta}y_t\right).
\end{align}

We also define the momentum ratio
\begin{align}
    x_t := \frac{m_{t-1}}{m_t}.
\end{align}
Starting from the first-moment recursion
\begin{align}
    m_t = \beta m_{t-1} + (1-\beta) g_t,
\end{align}
we solve for the current gradient:
\begin{align}
    g_t
    =
    \frac{m_t-\beta m_{t-1}}{1-\beta}
    =
    \frac{m_t}{1-\beta}(1-\beta x_t).
    \label{eq:g_from_x_appendix}
\end{align}
Substituting~\cref{eq:g_from_x_appendix} into the second-moment recursion
\begin{align}
    v_t = \beta v_{t-1} + (1-\beta) g_t^2
\end{align}
gives
\begin{align}
    v_t
    =
    \beta v_{t-1}
    +
    \frac{m_t^2}{1-\beta}(1-\beta x_t)^2.
    \label{eq:v_recursion_x_appendix}
\end{align}

Next, express both \(v_t\) and \(v_{t-1}\) through the transformed state:
\begin{align}
    v_t = m_t^2\left(1+\frac{\beta}{1-\beta}y_t\right),
    \qquad
    v_{t-1} = m_{t-1}^2\left(1+\frac{\beta}{1-\beta}y_{t-1}\right).
\end{align}
Using \(m_{t-1}=x_t m_t\), \cref{eq:v_recursion_x_appendix} becomes
\begin{align}
    m_t^2\left(1+\frac{\beta}{1-\beta}y_t\right)
    =
    \beta x_t^2 m_t^2\left(1+\frac{\beta}{1-\beta}y_{t-1}\right)
    +
    \frac{m_t^2}{1-\beta}(1-\beta x_t)^2.
\end{align}
Dividing by \(m_t^2\) yields
\begin{align}
    1+\frac{\beta}{1-\beta}y_t
    =
    \beta x_t^2\left(1+\frac{\beta}{1-\beta}y_{t-1}\right)
    +
    \frac{(1-\beta x_t)^2}{1-\beta}.
\end{align}
Multiplying both sides by \(1-\beta\) and simplifying gives
\begin{align}
    (1-\beta)+\beta y_t
    =
    \beta(1-\beta)x_t^2 + \beta^2 x_t^2 y_{t-1} + (1-\beta x_t)^2,
\end{align}
and therefore
\begin{align}
    y_t
    =
    \beta x_t^2 y_{t-1} + (1-x_t)^2.
    \label{eq:y_recursion_appendix}
\end{align}
This is the scalar recursion stated in the main text. Algorithm~\ref{alg:tr-adam} gives the detailed pseudocode for TR-Adam, including the quantized ratio-state storage and bias-corrected parameter update.

\paragraph{Numerical representation.}
Algorithm~\ref{alg:tr-adam} describes the FP32-momentum variant. The experiments emulate low-precision states in FP32 tensors rather than using packed storage. The current ratio is quantized before the parameter update. In the FP8-momentum variant, the fresh momentum working value is used for the current update, then quantized to FP8 E4M3 using a per-tensor absolute-maximum scale for state writeback and use in the next step. Thus the comparisons evaluate numerical effects of state quantization; they do not measure memory or throughput gains from a packed implementation.

\begin{algorithm}[t]
\caption{TR-Adam with quantized transformed ratio}
\label{alg:tr-adam}
\begin{algorithmic}[1]
\Require Parameters $\theta_0$, objective $f_t(\theta)$, learning rate $\eta$, tied momentum coefficient $\beta$, numerical constant $\epsilon$, weight decay $\lambda$, ratio-state codebook $\mathcal{Q}$.
\State Initialize $m_0 \leftarrow 0$ for all parameters.
\For{$t=1,\ldots,T$}
    \State $g_t \leftarrow \nabla f_t(\theta_{t-1})$
    \State $m_t \leftarrow \beta m_{t-1} + (1-\beta) g_t$
    \If{$t=1$}
        \State $\widetilde y_t \leftarrow 1$ \Comment{initial transformed ratio }
    \Else
        \State $x_t \leftarrow m_{t-1}/m_t$
        \State $\widetilde y_t \leftarrow \beta x_t^2 y_{t-1} + (1-x_t)^2$
    \EndIf
    \If{$\widetilde y_t$ is non-finite or $\widetilde y_t>\max\mathcal Q$}
        \State $\widetilde y_t\leftarrow\max\mathcal Q$ \Comment{saturate before quantization}
    \EndIf
    \State $y_t \leftarrow Q_{\mathcal{Q}}(\widetilde y_t)$ \Comment{nearest codebook value, FP4-style quantization}
    \State $\widehat m_t \leftarrow m_t/(1-\beta^t)$ \Comment{first-moment bias correction}
    \State $r_t \leftarrow \frac{\beta}{1-\beta} y_t$
    \State $d_t \leftarrow |m_t|\sqrt{\frac{1+r_t}{1-\beta^t}}+\epsilon$ \Comment{denominator bias correction}
    \State $\theta_{t-\frac{1}{2}} \leftarrow (1-\eta\lambda)\theta_{t-1}$ \Comment{decoupled weight decay}
    \State $\theta_t \leftarrow \theta_{t-\frac{1}{2}} - \eta \widehat m_t / d_t$
\EndFor
\end{algorithmic}
\end{algorithm}

\begin{figure*}[htb]
    \centering
    \includegraphics[width=\textwidth]{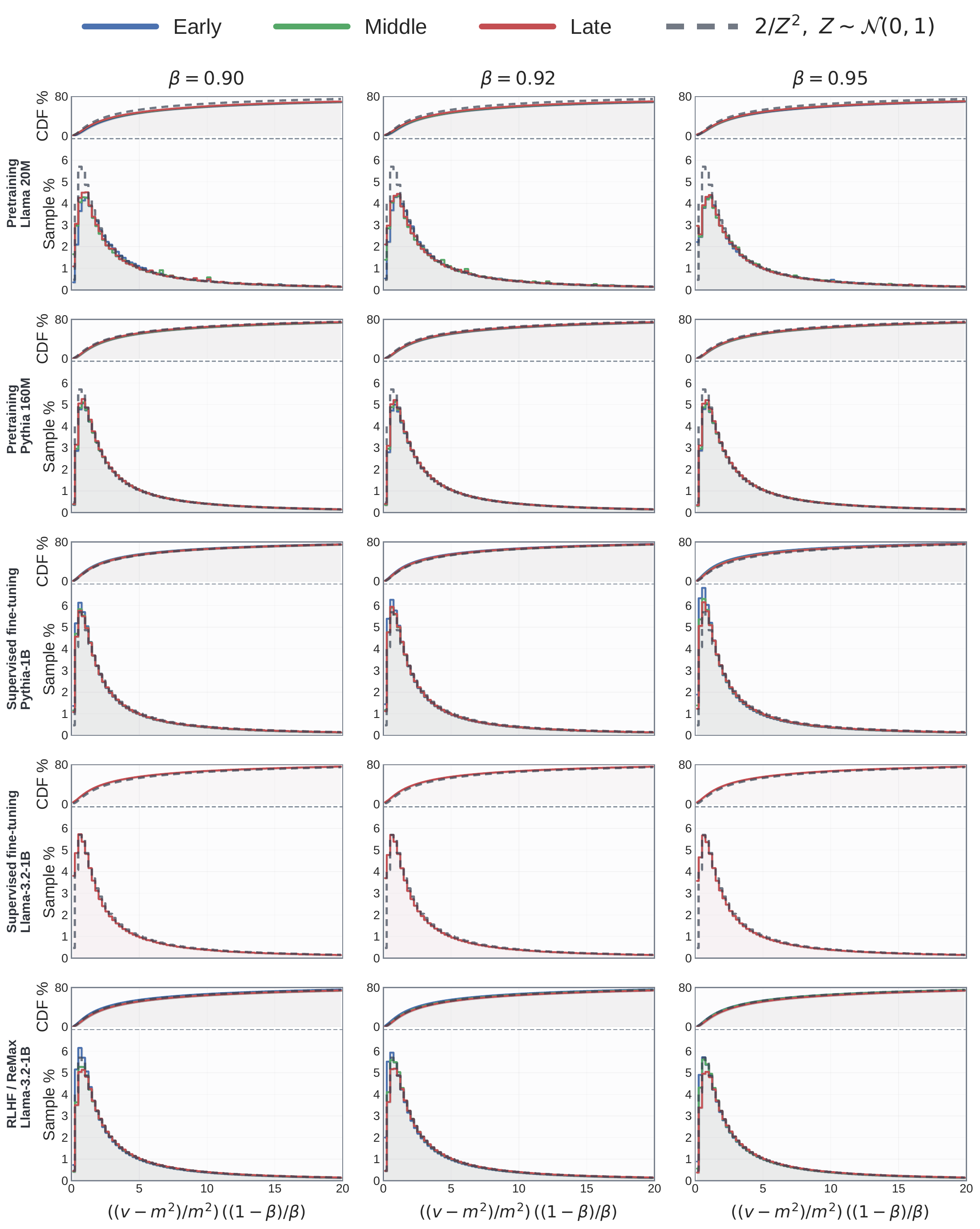}
    \caption{
    Distributions of the Adam-derived transformed variable \(y_t\) across five representative training settings: Llama-style 20M and Pythia-style 160M pre-training, Pythia-1B and Llama-3.2-1B supervised fine-tuning, and Llama-3.2-1B-Instruct ReMax training.
    Each block shows the empirical CDF at the top and the histogram at the bottom for different training stages.
    Across model families, scales, and training regimes, the transformed state exhibits a consistent heavy-tailed shape: a concentrated bulk near small values together with a long right tail.
    The dashed reference curve \(2/Z^2\), with \(Z\sim\mathcal{N}(0,1)\), is included to highlight the shared inverse-square-like structure.
    }
    \label{fig:five-model-y-dist}
\end{figure*}

\appendixsection{Additional Experimental Results}
\label{more_exp}
\subsection{Original language-model comparisons}
\label{app:original-results}

This section provides additional quantitative and distributional evidence for the robustness of the transformed ratio-state representation. 
\Cref{fig:five-model-y-dist} first shows that the transformed variable \(y_t\) has a broadly consistent bulk-and-tail distribution across five representative settings, including pre-training, SFT, and ReMax.
We then report final validation metrics in \cref{tab:pretrain-last3-val,tab:sft-last3-val,tab:remax-last3-val}, averaging over the final three validation points to reduce noise from individual checkpoints.

Across these settings, TR-Adam remains close to full Adam in pre-training and often improves over Adam in post-training, suggesting that the coarse transformed state preserves the main adaptive behavior while introducing a mild stabilizing effect.
\begin{table}[!htbp]
\centering
\small
\caption{Pre-training validation loss averaged over the final three validation points. For Llama-style 20M, the learning rate is selected from the full learning-rate sweep; for Pythia-style 160M, it is selected from $\{1,3,5,8,10\}\times10^{-3}$. Lower is better. Here $\Delta=\mathrm{TR\text{-}Adam}-\mathrm{Adam}$.}
\label{tab:pretrain-last3-val}
\begin{tabular}{llccccc}
\toprule
Model & $\beta$ & Adam & \makecell{TR-Adam\\(FP32-$m_t$, 4-bit $y_t$)} & \makecell{TR-Adam\\(FP8-$m_t$, 4-bit $y_t$)} & \makecell{$\Delta$\\FP32} & \makecell{$\Delta$\\FP8} \\
\midrule
Llama-style 20M & 0.90 & \textbf{3.694} & 3.706 & 3.710 & +0.012 & +0.016 \\
Llama-style 20M & 0.92 & \textbf{3.676} & 3.689 & 3.688 & +0.013 & +0.012 \\
Llama-style 20M & 0.95 & 3.672 & 3.672 & \textbf{3.670} & -0.000 & -0.002 \\
Llama-style 20M & 0.97 & 3.692 & 3.677 & \textbf{3.675} & -0.014 & -0.016 \\
Pythia-style 160M & 0.90 & \textbf{3.373} & 3.378 & 3.387 & +0.004 & +0.013 \\
Pythia-style 160M & 0.92 & \textbf{3.355} & 3.365 & 3.369 & +0.010 & +0.014 \\
Pythia-style 160M & 0.95 & \textbf{3.342} & 3.347 & 3.346 & +0.005 & +0.004 \\
Pythia-1B & 0.90 & 2.985 & 2.979 & \textbf{2.979} & -0.006 & -0.007 \\
Pythia-1B & 0.92 & 2.996 & \textbf{2.977} & 2.980 & -0.019 & -0.016 \\
Pythia-1B & 0.95 & 2.982 & \textbf{2.973} & 2.974 & -0.009 & -0.008 \\

\bottomrule
\end{tabular}
\end{table}

\begin{table}[!htbp]
\centering
\small
\caption{SFT validation loss averaged over the final three validation points. Lower is better. Here $\Delta=\mathrm{TR\text{-}Adam}-\mathrm{Adam}$.}
\label{tab:sft-last3-val}
\begin{tabular}{llccccc}
\toprule
Model & $\beta$ & Adam & \makecell{TR-Adam\\(FP32-$m_t$, 4-bit $y_t$)} & \makecell{TR-Adam\\(FP8-$m_t$, 4-bit $y_t$)} & \makecell{$\Delta$\\FP32} & \makecell{$\Delta$\\FP8} \\
\midrule
Pythia-1B & 0.90 & 1.299 & \textbf{1.297} & 1.297 & -0.002 & -0.002 \\
Pythia-1B & 0.92 & 1.302 & \textbf{1.295} & 1.295 & -0.007 & -0.006 \\
Pythia-1B & 0.95 & 1.309 & \textbf{1.299} & 1.299 & -0.011 & -0.011 \\
Llama-3.2-1B & 0.90 & 1.075 & \textbf{1.072} & 1.072 & -0.002 & -0.002 \\
Llama-3.2-1B & 0.92 & 1.076 & \textbf{1.071} & 1.071 & -0.005 & -0.005 \\
Llama-3.2-1B & 0.95 & 1.082 & \textbf{1.073} & 1.073 & -0.009 & -0.009 \\
\bottomrule
\end{tabular}
\end{table}

\clearpage
\begin{table}[!htbp]
\centering
\small
\caption{ReMax validation reward averaged over the final three validation points. Higher is better. Here $\Delta=\mathrm{TR\text{-}Adam}-\mathrm{Adam}$.}
\label{tab:remax-last3-val}
\begin{tabular}{lccccc}
\toprule
$\beta$ & Adam & \makecell{TR-Adam\\(FP32-$m_t$, 4-bit $y_t$)} & \makecell{TR-Adam\\(FP8-$m_t$, 4-bit $y_t$)} & \makecell{$\Delta$\\FP32} & \makecell{$\Delta$\\FP8} \\
\midrule
0.90 & 0.788 & \textbf{0.825} & 0.813 & +0.036 & +0.025 \\
0.92 & 0.800 & \textbf{0.824} & 0.819 & +0.024 & +0.019 \\
0.95 & 0.808 & \textbf{0.817} & 0.811 & +0.009 & +0.003 \\
\bottomrule
\end{tabular}
\end{table}

\input{additional_experiments}

\appendixsection{Datasets and Implementation Details}
\label{app:data_impl}
Across all optimizer comparisons, we keep the dataset, model architecture, batch size, learning-rate schedule, and evaluation protocol fixed, and only change the optimizer update. 
For short-run supervised experiments, including \texttt{plainLM}\footnote{\url{https://github.com/Niccolo-Ajroldi/plainLM}} pre-training ablations and \texttt{VERL}\footnote{\url{https://github.com/volcengine/verl}} SFT, we use zero weight decay to isolate optimizer-state effects and provide a cleaner comparison with Signum-style updates. 
For long-run supervised experiments, such as the 20B-token setting, we use weight decay \(0.1\). 
For SFT and RLHF, we follow the default \texttt{VERL} learning-rate settings, \(1\times10^{-5}\) and \(1\times10^{-6}\), respectively; for RLHF, we also keep the default actor weight decay \(0.01\).

For learning-rate selection, SFT and RLHF use the default \texttt{VERL} learning rates described above. 
For pre-training, we tune learning rates separately at the smaller scales before reporting the best result for each method. 
For the Llama-style 20M experiments, we sweep the learning-rate grid shown in Fig.~\ref{fig:llama20m_lr_sweep_fp32_fp8} and Fig.~\ref{fig:signum_adam_lr_sweeps}. 
For the Pythia-style 160M experiments, we sweep \(\{1,3,5,8,10\}\times 10^{-3}\) and select the best learning rate for each optimizer variant and \(\beta\). 
For the Pythia-1B 20B-token experiment, a full sweep is too expensive; we therefore run a short 5B-token sweep for Adam at \(\beta=0.95\) over \(\{0.001,0.003,0.005,0.008\}\), select \(0.001\), and use this learning rate for all \(\beta\) values and optimizer variants.

For the small-scale pre-training models, we use compact decoder-only Transformer architectures implemented in \texttt{plainLM}. 
The Llama-style 20M model is our own \texttt{plainLM} implementation based on Hugging Face's \texttt{LlamaForCausalLM}/\texttt{LlamaConfig} interface, with its scale and layout adapted from a community 20M Llama-style checkpoint\footnote{\url{https://huggingface.co/alan918727/TinyStories-LLaMA2-20M-256h-4l-colab}}; we do not use its pretrained weights. 
The Pythia-style 160M model follows a Pythia-like architecture~\citep{biderman2023pythia} constructed in the same codebase. 
For the 1B-scale experiments, we use the official Llama-3.2-1B and Pythia-1B architectures; in pre-training experiments, all models are trained from scratch by reinitializing the parameters.

\paragraph{Pre-training.}
For pre-training, we use subsets of the FineWeb corpus~\citep{penedo2024fineweb} and implement all experiments in \texttt{plainLM}.

\begin{table}[!htb]
\centering

\caption{Pre-training experimental setup.}
\label{tab:pretrain-setup}
\begin{tabular}{ll}
\toprule
Dataset & FineWeb 10B / 10B / 100B \\
Models & Llama-style 20M / Pythia-style 160M / Pythia-1B \\
Training budget & 1.5B / 4B / 20B tokens \\
Sequence length & 2048 \\
Batch size & 256 \\
Warmup & 512 / 1024 / 1024 steps \\
LR schedule & constant after warmup \\
Implementation & \texttt{plainLM} \\
\bottomrule
\end{tabular}
\end{table}

\paragraph{Supervised Fine-Tuning (SFT).}
For SFT, we use UltraChat~\citep{ding2023ultrachat}, formatted as multi-turn conversations and trained with the standard teacher-forcing objective. We fine-tune 1B-scale backbones, including Pythia-1B and Llama-3.2-1B, using the \texttt{VERL} SFT trainer.

\begin{table}[!htb]
\centering
\caption{SFT experimental setup.}
\label{tab:sft-setup}
\begin{tabular}{ll}
\toprule
Dataset & UltraChat \\
Models & Pythia-1B, Llama-3.2-1B \\
Training budget & 2000 steps \\
Maximum sequence length & 2048 \\
Batch size & 128 \\
Learning rate & \(1\times10^{-5}\) \\
LR schedule & constant after warmup \\
Implementation & \texttt{VERL} SFT trainer \\
\bottomrule
\end{tabular}
\end{table}

\paragraph{RLHF.}
For RLHF, we use UltraFeedback~\citep{cui2023ultrafeedback} as the prompt and preference source. We use Llama-3.2-1B-Instruct as the actor model and Skywork-Reward-V2-Llama-3.2-3B~\citep{liu2025skyworkrewardv2} as the reward model. The RL training pipeline is implemented with \texttt{VERL}, with our optimizer integrated into the actor update.

\begin{table}[!htb]
\centering
\caption{RLHF/ReMax experimental setup.}
\label{tab:rlhf-setup}
\begin{tabular}{ll}
\toprule
Dataset & UltraFeedback \\
Actor model & Llama-3.2-1B-Instruct \\
Reward model & Skywork-Reward-V2-Llama-3.2-3B \\
Algorithm & ReMax~\cite{li2024remax} \\
Training budget & 600 steps \\
Batch size & 64 prompts \\
Maximum prompt length & 768 \\
Maximum response length & 1024 \\
Rollout temperature & 1.0 \\
Actor learning rate & \(1\times10^{-6}\) \\
Actor LR schedule & constant, no warmup \\
Implementation & \texttt{VERL} RL trainer \\
\bottomrule
\end{tabular}
\end{table}

The original 1B-scale experiments, including pre-training, SFT, and RLHF, are run on \(8\times\)H200 GPUs, while the original smaller-scale experiments are run on \(16\times\)RTX 4090 GPUs.

\appendixsection{Broader Impact}
This work primarily aims to improve our understanding of Adam's internal adaptive dynamics. By identifying a natural-scale ratio-state representation in the tied-\(\beta\) regime, it provides a more transparent view of how the mean estimate and adaptive denominator interact during training. This mechanistic perspective may help guide optimizer analysis, hyperparameter choices, and the design of simpler Adam-like methods. This paper is methodological and does not involve data from human subjects, user studies, or deployment of a user-facing system. We do not directly study downstream model behavior; our focus is on optimizer dynamics and training efficiency.

A future packed implementation of the low-precision representation could reduce optimizer-state memory and make training more accessible; the present experiments do not measure such savings. Lower training costs could support beneficial applications but could also lower barriers to harmful downstream uses. These effects depend on the trained models and deployment context, which are outside our experimental scope.

%% file: additional_experiments.tex
\subsection{Reproducibility across random seeds}
\label{app:multiseed}
We repeat the Llama-style 20M pre-training and Llama-3.2-1B-Instruct ReMax RLHF comparisons using three random seeds. Tables~\ref{tab:multiseed-pretrain} and~\ref{tab:multiseed-rlhf} retain the original single-seed results alongside the repeated experiments. For three-seed rows, we report the mean and standard deviation across runs, rounded to three decimal places; these are standard deviations, not standard errors or confidence intervals. The FP32 and FP8 variants refer to momentum precision, with a 4-bit transformed-ratio state in both cases. Each pre-training score averages validation loss at steps 2,400, 2,600, and 2,800; each ReMax score averages validation reward at steps 550, 575, and 600.

The repeated results preserve the main qualitative trends. On Llama-style 20M, TR-Adam has slightly higher loss at $\beta=0.90,0.92$ and slightly lower loss at $\beta=0.95,0.97$. In the RLHF comparison, both TR-Adam variants have higher mean reward than Adam at the three tested decay rates. These small samples characterize run-to-run variation in selected settings; they do not establish statistical significance or replace repeated evaluation of every experiment.

\begin{table}[htbp]
\centering
\small
\caption{Llama-style 20M pre-training validation loss; lower is better. Single-seed entries are the original results; three-seed entries are mean $\pm$ standard deviation.}
\label{tab:multiseed-pretrain}
\begin{tabular}{llccc}
\toprule
$\beta$ & Runs & Adam & \makecell{TR-Adam\\FP32 momentum} & \makecell{TR-Adam\\FP8 momentum} \\
\midrule
$0.90$ & 1 seed & 3.694 & 3.706 & 3.710 \\
$0.90$ & 3 seeds & $3.697\pm0.012$ & $3.711\pm0.004$ & $3.711\pm0.003$ \\
$0.92$ & 1 seed & 3.676 & 3.689 & 3.688 \\
$0.92$ & 3 seeds & $3.676\pm0.002$ & $3.691\pm0.004$ & $3.694\pm0.006$ \\
$0.95$ & 1 seed & 3.672 & 3.672 & 3.670 \\
$0.95$ & 3 seeds & $3.673\pm0.002$ & $3.671\pm0.002$ & $3.670\pm0.002$ \\
$0.97$ & 1 seed & 3.692 & 3.677 & 3.675 \\
$0.97$ & 3 seeds & $3.693\pm0.005$ & $3.672\pm0.005$ & $3.676\pm0.003$ \\
\bottomrule
\end{tabular}
\end{table}

\begin{table}[htbp]
\centering
\small
\caption{Llama-3.2-1B-Instruct ReMax RLHF evaluation reward; higher is better. Single-seed entries are the original results; three-seed entries are mean $\pm$ standard deviation.}
\label{tab:multiseed-rlhf}
\begin{tabular}{llccc}
\toprule
$\beta$ & Runs & Adam & \makecell{TR-Adam\\FP32 momentum} & \makecell{TR-Adam\\FP8 momentum} \\
\midrule
$0.90$ & 1 seed & 0.788 & 0.825 & 0.813 \\
$0.90$ & 3 seeds & $0.802\pm0.013$ & $0.811\pm0.017$ & $0.818\pm0.004$ \\
$0.92$ & 1 seed & 0.800 & 0.824 & 0.819 \\
$0.92$ & 3 seeds & $0.798\pm0.002$ & $0.820\pm0.005$ & $0.818\pm0.005$ \\
$0.95$ & 1 seed & 0.808 & 0.817 & 0.811 \\
$0.95$ & 3 seeds & $0.801\pm0.015$ & $0.835\pm0.019$ & $0.805\pm0.009$ \\
\bottomrule
\end{tabular}
\end{table}

\clearpage
\subsection{Distributional structure across architectures and model scales}
\label{app:architecture-scale}
\subsubsection{Non-Transformer performance and training-stage distributions}
We evaluate two multilayer perceptrons (MLPs; 4.73M and 427.7K parameters) on CIFAR-10 and an 11.22M-parameter ResNet-18~\citep{he2015deepresidual} on CIFAR-100~\citep{krizhevsky2009learning}. Each model is trained for 3,000 optimizer steps with learning rate $2\times10^{-4}$ and batch size 512, $\epsilon=10^{-8}$, and no weight decay. TR-Adam uses FP32 momentum and the 16-value ratio codebook in Algorithm~\ref{alg:tr-adam}. In these experiments the quantized ratio values are represented in FP32 tensors; the 4-bit description refers to the codebook cardinality, not measured packed-state memory. For each seed, test loss is averaged over evaluations at steps 2,842, 2,940, and 3,000. Table~\ref{tab:cifar-performance} reports the mean and standard deviation of this statistic over seeds 1 and 2. A displayed standard deviation of $0.000$ reflects rounding to three decimal places, not necessarily zero variation.

TR-Adam gives losses close to Adam across all nine settings. The differences are small and not uniformly favorable: for example, ResNet-18 at $\beta=0.90$ has a slightly higher mean loss with TR-Adam. The evidence supports applicability beyond Transformers under the tested budget, without establishing a universal performance advantage.

\begin{table}[htbp]
\centering
\small
\caption{Non-Transformer test loss (mean $\pm$ standard deviation over two seeds; lower is better).}
\label{tab:cifar-performance}
\begin{tabular}{lccc}
\toprule
Model / dataset & $\beta$ & Adam & TR-Adam \\
\midrule
MLP 4.73M / CIFAR-10 & 0.90 & $1.121\pm0.005$ & $1.119\pm0.006$ \\
MLP 4.73M / CIFAR-10 & 0.92 & $1.124\pm0.003$ & $1.122\pm0.000$ \\
MLP 4.73M / CIFAR-10 & 0.95 & $1.128\pm0.003$ & $1.117\pm0.008$ \\
MLP 427.7K / CIFAR-10 & 0.90 & $1.289\pm0.003$ & $1.280\pm0.010$ \\
MLP 427.7K / CIFAR-10 & 0.92 & $1.290\pm0.004$ & $1.276\pm0.007$ \\
MLP 427.7K / CIFAR-10 & 0.95 & $1.298\pm0.000$ & $1.275\pm0.000$ \\
ResNet-18 / CIFAR-100 & 0.90 & $1.576\pm0.009$ & $1.588\pm0.026$ \\
ResNet-18 / CIFAR-100 & 0.92 & $1.569\pm0.001$ & $1.568\pm0.025$ \\
ResNet-18 / CIFAR-100 & 0.95 & $1.557\pm0.012$ & $1.545\pm0.008$ \\
\bottomrule
\end{tabular}
\end{table}

Using Adam moment states and seed 1, we separately measure the empirical distribution $F(c)=\Pr(y_t\leq c)$ during early (steps 1--1,000), middle (1,001--2,000), and late (2,001--3,000) training, pooling all parameter entries at every optimizer step within each interval. Tables~\ref{tab:cdf-mlp-large}--\ref{tab:cdf-resnet} compare the measured CDF values with the zero-shift reference $2/Z^2$, where $Z\sim\mathcal N(0,1)$. The entries are probabilities, not percentages. These distributional measurements and the optimizer-performance comparisons answer distinct questions: similarity of the observed ratio distribution does not by itself establish the accuracy of a recursively quantized optimizer.

The later-stage CDFs remain close across the tested decay rates. The smaller MLP exhibits a clearer early-stage shift, while its middle and late distributions are similar. The reference is an approximate description: systematic differences remain, and the tabulated thresholds do not constitute a goodness-of-fit test or characterize the full tail.
\clearpage

\begin{table}[htbp]
\centering
\small
\caption{Training-stage hidden-ratio CDF for the 4.73M-parameter MLP on CIFAR-10.}
\label{tab:cdf-mlp-large}
\begin{tabular}{llccccccc}
\toprule
$\beta$ & Stage & $F(1)$ & $F(2)$ & $F(4)$ & $F(7)$ & $F(10)$ & $F(15)$ & $F(20)$ \\
\midrule
-- & $2/Z^2$ reference & 0.157 & 0.317 & 0.480 & 0.593 & 0.655 & 0.715 & 0.752 \\
0.90 & Early & 0.142 & 0.295 & 0.456 & 0.570 & 0.633 & 0.695 & 0.733 \\
0.90 & Middle & 0.143 & 0.297 & 0.460 & 0.575 & 0.639 & 0.701 & 0.739 \\
0.90 & Late & 0.148 & 0.303 & 0.466 & 0.581 & 0.644 & 0.705 & 0.743 \\
0.92 & Early & 0.143 & 0.296 & 0.456 & 0.571 & 0.633 & 0.695 & 0.732 \\
0.92 & Middle & 0.143 & 0.296 & 0.458 & 0.574 & 0.638 & 0.700 & 0.738 \\
0.92 & Late & 0.148 & 0.302 & 0.464 & 0.579 & 0.642 & 0.704 & 0.742 \\
0.95 & Early & 0.145 & 0.297 & 0.455 & 0.568 & 0.629 & 0.689 & 0.726 \\
0.95 & Middle & 0.143 & 0.294 & 0.455 & 0.570 & 0.633 & 0.695 & 0.734 \\
0.95 & Late & 0.150 & 0.301 & 0.462 & 0.577 & 0.640 & 0.702 & 0.740 \\
\bottomrule
\end{tabular}
\end{table}

\begin{table}[htbp]
\centering
\small
\caption{Training-stage hidden-ratio CDF for the 427.7K-parameter MLP on CIFAR-10.}
\label{tab:cdf-mlp-small}
\begin{tabular}{llccccccc}
\toprule
$\beta$ & Stage & $F(1)$ & $F(2)$ & $F(4)$ & $F(7)$ & $F(10)$ & $F(15)$ & $F(20)$ \\
\midrule
-- & $2/Z^2$ reference & 0.157 & 0.317 & 0.480 & 0.593 & 0.655 & 0.715 & 0.752 \\
0.90 & Early & 0.184 & 0.344 & 0.501 & 0.610 & 0.670 & 0.727 & 0.762 \\
0.90 & Middle & 0.143 & 0.302 & 0.466 & 0.582 & 0.645 & 0.707 & 0.745 \\
0.90 & Late & 0.141 & 0.300 & 0.464 & 0.580 & 0.644 & 0.706 & 0.744 \\
0.92 & Early & 0.190 & 0.349 & 0.506 & 0.614 & 0.672 & 0.729 & 0.764 \\
0.92 & Middle & 0.142 & 0.300 & 0.464 & 0.580 & 0.643 & 0.705 & 0.743 \\
0.92 & Late & 0.140 & 0.298 & 0.463 & 0.579 & 0.643 & 0.705 & 0.743 \\
0.95 & Early & 0.209 & 0.367 & 0.520 & 0.626 & 0.683 & 0.738 & 0.772 \\
0.95 & Middle & 0.141 & 0.300 & 0.464 & 0.580 & 0.643 & 0.705 & 0.743 \\
0.95 & Late & 0.139 & 0.296 & 0.461 & 0.577 & 0.641 & 0.703 & 0.742 \\
\bottomrule
\end{tabular}
\end{table}

\clearpage

\begin{table}[htbp]
\centering
\small
\caption{Training-stage hidden-ratio CDF for the 11.22M-parameter ResNet-18 on CIFAR-100.}
\label{tab:cdf-resnet}
\begin{tabular}{llccccccc}
\toprule
$\beta$ & Stage & $F(1)$ & $F(2)$ & $F(4)$ & $F(7)$ & $F(10)$ & $F(15)$ & $F(20)$ \\
\midrule
-- & $2/Z^2$ reference & 0.157 & 0.317 & 0.480 & 0.593 & 0.655 & 0.715 & 0.752 \\
0.90 & Early & 0.157 & 0.317 & 0.480 & 0.593 & 0.655 & 0.715 & 0.752 \\
0.90 & Middle & 0.147 & 0.306 & 0.470 & 0.585 & 0.648 & 0.709 & 0.747 \\
0.90 & Late & 0.148 & 0.308 & 0.471 & 0.586 & 0.649 & 0.710 & 0.748 \\
0.92 & Early & 0.159 & 0.319 & 0.481 & 0.594 & 0.656 & 0.716 & 0.753 \\
0.92 & Middle & 0.144 & 0.303 & 0.468 & 0.583 & 0.646 & 0.708 & 0.746 \\
0.92 & Late & 0.145 & 0.304 & 0.468 & 0.584 & 0.647 & 0.708 & 0.746 \\
0.95 & Early & 0.165 & 0.326 & 0.486 & 0.599 & 0.660 & 0.719 & 0.755 \\
0.95 & Middle & 0.140 & 0.298 & 0.462 & 0.578 & 0.642 & 0.704 & 0.743 \\
0.95 & Late & 0.141 & 0.299 & 0.463 & 0.579 & 0.643 & 0.705 & 0.743 \\
\bottomrule
\end{tabular}
\end{table}

\subsubsection{An early-stage distributional check at 8B scale}
Table~\ref{tab:cdf-8b} reports early-stage measurements from Llama-3.1-8B-Instruct under ReMax RLHF. Here ``Early'' means the first third of a 600-step budget: the statistics pool snapshots at steps 10, 20, $\ldots$, 200 from both FSDP ranks, weighted by their finite-entry counts. The runs use seed 42, tied-$\beta$ AdamW, a prompt batch size of 64, and maximum prompt/response lengths of 768/1,024 tokens. This is a step-based interval; variable response lengths do not define a fixed token count. At the reported thresholds, the CDFs for $\beta=0.90$ and $0.95$ are close to the $2/Z^2$ reference. This extends the distributional observation to an 8B model, but the available measurements cover only the early stage. They do not establish stability throughout training, nor do they constitute an 8B comparison of TR-Adam performance, memory consumption, or runtime.

\begin{table}[htbp]
\centering
\small
\caption{Early-stage hidden-ratio CDF for Llama-3.1-8B-Instruct under ReMax RLHF, pooling snapshots every 10 steps through step 200.}
\label{tab:cdf-8b}
\begin{tabular}{llccccccc}
\toprule
$\beta$ & Stage & $F(1)$ & $F(2)$ & $F(4)$ & $F(7)$ & $F(10)$ & $F(15)$ & $F(20)$ \\
\midrule
-- & $2/Z^2$ reference & 0.157 & 0.317 & 0.480 & 0.593 & 0.655 & 0.715 & 0.752 \\
0.90 & Early & 0.150 & 0.315 & 0.478 & 0.591 & 0.651 & 0.710 & 0.747 \\
0.95 & Early & 0.168 & 0.334 & 0.495 & 0.605 & 0.664 & 0.722 & 0.758 \\
\bottomrule
\end{tabular}
\end{table}

\clearpage
\subsection{Ratio distributions at large decay rates}
\label{app:large-beta}
We examine the distribution of the exact Adam-derived ratio at tied decay rates $\beta=0.99$ and $0.999$ on Llama-style 20M.

Table~\ref{tab:large-beta-cdf} measures the exact Adam-derived ratio on Llama-style 20M, using learning rates $0.004$ for $\beta=0.99$ and $0.0015$ for $\beta=0.999$. The first three intervals pool every optimizer step and parameter entry: steps 1--954, 955--1,908, and 1,909--2,861, with 524,288 tokens per step (approximately 0--0.5B, 0.5B--1B, and 1B--1.5B tokens). CDFs are normalized by finite ratio entries; undefined or non-finite ratios are excluded (about $1.1$--$3.4\%$ across these stages). At $\beta=0.99$, the early interval differs from the reference, while the later intervals are closer. At $\beta=0.999$, the CDF is substantially shifted toward smaller values throughout the first 1.5B tokens; for example, $F(1)$ decreases from $0.686$ to $0.476$, compared with the reference value $0.157$. The extended run gives $F(1)=0.213$ in the 10B--12B interval. This row pools 38 snapshots at steps 19,100--22,800, spaced 100 steps apart (10.014B--11.954B tokens); $1.036\%$ of ratio entries are non-finite and excluded.

This evolution is consistent with the shifted reference $Y_c=2/(Z+c)^2$ and a changing local mean-to-noise ratio. The CDFs alone do not identify the cause of the shift or prove convergence to the zero-shift law. They show that approximate distributional stability should not be assumed uniformly across decay rates and training stages. The extended run provides distributional measurements, not a new optimizer-performance comparison.

\begin{table}[htbp]
\centering
\small
\caption{Hidden-ratio CDF at large tied decay rates, including the 10B--12B interval of extended training at $\beta=0.999$.}
\label{tab:large-beta-cdf}
\begin{tabular}{llcccccccc}
\toprule
$\beta$ & Token interval & $F(0)$ & $F(1)$ & $F(2)$ & $F(4)$ & $F(7)$ & $F(10)$ & $F(15)$ & $F(20)$ \\
\midrule
$2/Z^2$ & Reference & 0 & 0.157 & 0.317 & 0.480 & 0.593 & 0.655 & 0.715 & 0.752 \\
0.99 & 0--0.5B & 0 & 0.319 & 0.457 & 0.582 & 0.668 & 0.715 & 0.763 & 0.792 \\
0.99 & 0.5B--1.0B & 0 & 0.158 & 0.313 & 0.467 & 0.574 & 0.633 & 0.692 & 0.728 \\
0.99 & 1.0B--1.5B & 0 & 0.152 & 0.309 & 0.467 & 0.578 & 0.638 & 0.698 & 0.734 \\
0.999 & 0--0.5B & 0 & 0.686 & 0.792 & 0.853 & 0.889 & 0.907 & 0.924 & 0.934 \\
0.999 & 0.5B--1.0B & 0 & 0.561 & 0.677 & 0.769 & 0.824 & 0.852 & 0.878 & 0.895 \\
0.999 & 1.0B--1.5B & 0 & 0.476 & 0.599 & 0.704 & 0.773 & 0.809 & 0.843 & 0.863 \\
0.999 & 10B--12B & 0 & 0.213 & 0.372 & 0.524 & 0.629 & 0.686 & 0.741 & 0.774 \\
\bottomrule
\end{tabular}
\end{table}

\clearpage
\subsection{Unequal decay rates: distributional changes and scope}
\label{app:unequal-beta-exp}
Appendix~\ref{app:balanced_regime_motivation} explains why tied decay rates are structurally special: the nonnegative ratio-state property holds for all gradient histories in the tied setting, but is not guaranteed for unequal decays. To examine the distribution empirically, define
\[
\bar\beta=\frac{\beta_1+\beta_2}{2},\qquad
r_t^{\mathrm u}=\frac{v_t^{\mathrm u}}{(m_t^{\mathrm u})^2}-1,\qquad
\widetilde y_t^{\mathrm u}=\frac{1-\bar\beta}{\bar\beta}r_t^{\mathrm u}.
\]
The measurements use Pythia-style 160M. Here $(m_t^{\mathrm u},v_t^{\mathrm u})$ are the exact Adam moments with the specified unequal decays; the ratio is defined where $m_t^{\mathrm u}\ne0$. We measure $F(c)=\Pr(\widetilde y_t^{\mathrm u}\leq c)$, rather than applying the tied-$\beta$ TR-Adam recursion to unequal decays. The unequal-decay rows pool all parameter entries at steps 1--2,000; the tied baseline is the previously recorded Early snapshot at step 2,000. These aggregation windows differ, so the comparison is descriptive rather than a matched isolation of the decay-rate effect. The histogram column $F(0^-)$ counts strictly negative finite ratios; other columns include this mass and the histogram bins below the indicated threshold.

For context, let $(\bar m_t,\bar v_t)$ be moments computed from the same gradient history using tied decay $\bar\beta$. Whenever the relative deviations are defined,
\[
\varepsilon_{m,t}=\frac{m_t^{\mathrm u}-\bar m_t}{\bar m_t},\qquad
\varepsilon_{v,t}=\frac{v_t^{\mathrm u}-\bar v_t}{\bar v_t},\qquad
1+r_t^{\mathrm u}=(1+\bar r_t)
\frac{1+\varepsilon_{v,t}}{(1+\varepsilon_{m,t})^2}.
\]
This identity shows why small relative moment changes can preserve the ratio away from small-momentum events, while closeness of decay rates alone does not guarantee a uniformly small perturbation.

\begin{table}[htbp]
\centering
\small
\caption{Pythia-style 160M CDF of the exact Adam-derived transformed ratio for tied and unequal decay rates. $F(0^-)$ measures strictly negative mass.}
\label{tab:unequal-beta-cdf}
\begin{tabular}{lcccccccc}
\toprule
$(\beta_1,\beta_2)$ & $F(0^-)$ & $F(1)$ & $F(2)$ & $F(4)$ & $F(7)$ & $F(10)$ & $F(15)$ & $F(20)$ \\
\midrule
$2/Z^2$ & 0 & 0.157 & 0.317 & 0.480 & 0.593 & 0.655 & 0.715 & 0.752 \\
$(.90,.90)$ & 0 & .131 & .290 & .456 & .573 & .637 & .699 & .738 \\
$(0.90,0.92)$ & $6.56\times10^{-6}$ & 0.145 & 0.290 & 0.445 & 0.559 & 0.623 & 0.686 & 0.725 \\
$(0.90,0.95)$ & $1.71\times10^{-3}$ & 0.170 & 0.309 & 0.456 & 0.565 & 0.627 & 0.688 & 0.727 \\
$(0.90,0.999)$ & $3.66\times10^{-2}$ & 0.201 & 0.277 & 0.364 & 0.438 & 0.485 & 0.537 & 0.572 \\
\bottomrule
\end{tabular}
\end{table}

At $(\beta_1,\beta_2)=(0.90,0.92)$ and $(0.90,0.95)$, the positive-range CDF values remain broadly similar to the tied baseline, but negative mass appears. At $(0.90,0.999)$, $F(0^-)$ rises to $0.0366$ and the positive-range CDF differs much more strongly; for example, $F(20)=0.572$ versus $0.738$ for the tied $(0.90,0.90)$ baseline. These diagnostics delimit the tied-regime claims. They do not validate a low-bit unequal-decay optimizer or establish equivalence of the resulting training trajectories.
\clearpage
\subsection{Numerical stability and the role of epsilon}
\label{app:epsilon-ablation}
\subsubsection{Optimization sensitivity}
On Llama-style 20M with FP32 momentum and the 16-value ratio codebook, Table~\ref{tab:epsilon-loss} varies the numerical constant $\epsilon$ around the default $10^{-6}$. Each loss averages validation at steps 2,400, 2,600, and 2,800. Learning rates are held at their default-$\epsilon$ values: Adam/TR-Adam use $0.024/0.018$, $0.010/0.006$, and $0.005/0.005$ for $\beta=0.90,0.95,0.97$, respectively. The $10^{-6}$ column retains the original paper results; other cells use seed 100, except the matched $\epsilon=10^{-4},\beta=0.95$ pair, which uses seed 10,000. This is a single-run sensitivity check, not a fully seed-matched sweep. NaN denotes a non-finite outcome rather than a valid loss. Standard Adam produces NaNs at $\epsilon=0$ in all three tested decay settings. TR-Adam produces finite losses using the special $\epsilon=0$ implementation described below; directly substituting zero into the denominator form also failed. For positive values from $10^{-12}$ to $10^{-6}$, the losses change modestly, whereas $10^{-4}$ gives consistently worse losses for both optimizers. These observations concern the tested implementations and settings, and do not imply that $\epsilon$ can be removed safely in general.

\begin{table}[htbp]
\centering
\small
\caption{Validation loss under numerical-constant ablation. Lower is better; the default is $\epsilon=10^{-6}$.}
\label{tab:epsilon-loss}
\begin{tabular}{llccccc}
\toprule
Optimizer & $\beta$ & $10^{-6}$ & $0$ & $10^{-12}$ & $10^{-8}$ & $10^{-4}$ \\
\midrule
Adam & 0.90 & 3.694 & NaN & 3.700 & 3.702 & 3.791 \\
Adam & 0.95 & 3.672 & NaN & 3.673 & 3.678 & 3.737 \\
Adam & 0.97 & 3.692 & NaN & 3.697 & 3.693 & 3.795 \\
TR-Adam & 0.90 & 3.706 & 3.722 & 3.727 & 3.721 & 3.804 \\
TR-Adam & 0.95 & 3.672 & 3.683 & 3.683 & 3.678 & 3.764 \\
TR-Adam & 0.97 & 3.677 & 3.680 & 3.676 & 3.669 & 3.786 \\
\bottomrule
\end{tabular}
\end{table}

For $\epsilon=0$, the implementation cancels $|m_t|$ algebraically before evaluating the update. Writing $b_t=1-\beta^t$ and $r_t=\frac{\beta}{1-\beta}\widehat y_t$, it computes
\[
u_t=\frac{\operatorname{sign}(m_t)}{\sqrt{b_t(1+r_t)}},\qquad \operatorname{sign}(0)=0.
\]
This equals the bias-corrected denominator form when $m_t\ne0$ and defines a zero update at $m_t=0$, avoiding its explicit $0/0$. For positive $\epsilon$, the implementation retains Algorithm~\ref{alg:tr-adam}'s denominator form. A nonnegative quantized state ensures $\sqrt{1+r_t}\geq1$, but this bound alone does not resolve the zero-momentum case. The ratio recurrence separately saturates NaN or positive-infinite candidates at the maximum codebook value and clips the candidate before quantization. These are implementation choices, distinct from adding a positive constant to the denominator.

\subsubsection{Layer-wise second-moment magnitudes}
Tables~\ref{tab:epsilon-magnitude-default} and~\ref{tab:epsilon-magnitude-small} measure uncorrected Adam second moments on Llama-style 20M, with seed 100 and a 2,861-step budget. They pool every parameter entry at every optimizer step across Early (1--954), Middle (955--1,908), and Late (1,909--2,861). Each reports the arithmetic mean $\mathbb E[\sqrt{v_t}]$ and the percentage of entries for which $\sqrt{v_t}\leq\epsilon$. The means are on the original scale, not means of $\log_{10}\sqrt{v_t}$. Percentages displayed as $0.000\%$ are rounded observations.

The condition $\sqrt{v_t}\gg\epsilon$ is not uniform across layers. With $\epsilon=10^{-6}$, roughly $40$--$71\%$ of embedding entries and $38$--$63\%$ of LM-head entries are at or below $\epsilon$. The fractions are much smaller in the Transformer blocks and final normalization layer. Even with $\epsilon=10^{-8}$, non-negligible fractions of embedding and LM-head entries remain below the threshold. Thus, a small change in aggregate validation loss does not establish that $\epsilon$ is negligible for every coordinate. The measured sensitivity qualifies the epsilon-free theoretical interpretation rather than providing a uniform justification for dropping the stabilizer.
\clearpage

\begin{table}[htbp]
\centering
\small
\caption{Layer-wise $\sqrt{v_t}$ magnitudes with $\epsilon=10^{-6}$: arithmetic means and percentages at or below the stabilizer. Measurements are aggregated across training stages.}
\label{tab:epsilon-magnitude-default}
\begin{tabular}{lcccccc}
\toprule
Layer & \makecell{$\beta=0.90$\\Mean} & \makecell{$\beta=0.90$\\$\leq\epsilon$ (\%)} & \makecell{$\beta=0.95$\\Mean} & \makecell{$\beta=0.95$\\$\leq\epsilon$ (\%)} & \makecell{$\beta=0.97$\\Mean} & \makecell{$\beta=0.97$\\$\leq\epsilon$ (\%)} \\
\midrule
Embedding & $1.428\times10^{-6}$ & 70.829 & $2.464\times10^{-6}$ & 50.863 & $3.666\times10^{-6}$ & 40.316 \\
Block 0 & $2.496\times10^{-5}$ & 3.513 & $4.978\times10^{-5}$ & 0.143 & $8.711\times10^{-5}$ & 0.052 \\
Block 1 & $1.110\times10^{-5}$ & 0.415 & $1.981\times10^{-5}$ & 0.273 & $3.047\times10^{-5}$ & 0.397 \\
Block 2 & $1.301\times10^{-5}$ & 0.345 & $2.113\times10^{-5}$ & 0.352 & $3.113\times10^{-5}$ & 0.523 \\
Block 3 & $1.285\times10^{-5}$ & 0.260 & $2.434\times10^{-5}$ & 0.323 & $3.401\times10^{-5}$ & 0.435 \\
Final norm & $1.575\times10^{-3}$ & 0.000 & $5.875\times10^{-4}$ & 0.000 & $4.229\times10^{-4}$ & 0.001 \\
LM head & $2.372\times10^{-6}$ & 62.816 & $3.857\times10^{-6}$ & 46.659 & $5.542\times10^{-6}$ & 37.638 \\
\bottomrule
\end{tabular}
\end{table}

\begin{table}[htbp]
\centering
\small
\caption{Layer-wise $\sqrt{v_t}$ magnitudes with $\epsilon=10^{-8}$: arithmetic means and percentages at or below the stabilizer. Measurements are aggregated across training stages.}
\label{tab:epsilon-magnitude-small}
\begin{tabular}{lcccccc}
\toprule
Layer & \makecell{$\beta=0.90$\\Mean} & \makecell{$\beta=0.90$\\$\leq\epsilon$ (\%)} & \makecell{$\beta=0.95$\\Mean} & \makecell{$\beta=0.95$\\$\leq\epsilon$ (\%)} & \makecell{$\beta=0.97$\\Mean} & \makecell{$\beta=0.97$\\$\leq\epsilon$ (\%)} \\
\midrule
Embedding & $1.114\times10^{-6}$ & 20.784 & $2.040\times10^{-6}$ & 14.831 & $3.183\times10^{-6}$ & 11.174 \\
Block 0 & $2.420\times10^{-5}$ & 0.000 & $4.926\times10^{-5}$ & 0.000 & $8.524\times10^{-5}$ & 0.000 \\
Block 1 & $1.219\times10^{-5}$ & 0.000 & $2.260\times10^{-5}$ & 0.000 & $3.139\times10^{-5}$ & 0.000 \\
Block 2 & $1.206\times10^{-5}$ & 0.000 & $2.084\times10^{-5}$ & 0.000 & $3.314\times10^{-5}$ & 0.000 \\
Block 3 & $1.133\times10^{-5}$ & 0.000 & $2.131\times10^{-5}$ & 0.001 & $3.653\times10^{-5}$ & 0.001 \\
Final norm & $2.155\times10^{-3}$ & 0.000 & $6.438\times10^{-4}$ & 0.000 & $4.429\times10^{-4}$ & 0.000 \\
LM head & $2.067\times10^{-6}$ & 19.282 & $3.497\times10^{-6}$ & 12.924 & $5.179\times10^{-6}$ & 8.934 \\
\bottomrule
\end{tabular}
\end{table}

\clearpage

%% file: checklist.tex
\section*{NeurIPS Paper Checklist}

\begin{enumerate}

\item {\bf Claims}
    \item[] Question: Do the main claims made in the abstract and introduction accurately reflect the paper's contributions and scope?
    \item[] Answer: \answerYes{} %
    \item[] Justification: Abstract and introduction accurately reflect the paper's contributions and scope

\item {\bf Limitations}
    \item[] Question: Does the paper discuss the limitations of the work performed by the authors?
    \item[] Answer: \answerYes{} %
    \item[] Justification: Discussed in paper
    \item[] Guidelines:
    \begin{itemize}
        \item The answer \answerNA{} means that the paper has no limitations while the answer \answerNo{} means that the paper has limitations, but those are not discussed in the paper.
        \item The authors are encouraged to create a separate ``Limitations'' section in their paper.
        \item The paper should point out any strong assumptions and how robust the results are to violations of these assumptions (e.g., independence assumptions, noiseless settings, model well-specification, asymptotic approximations only holding locally). The authors should reflect on how these assumptions might be violated in practice and what the implications would be.
        \item The authors should reflect on the scope of the claims made, e.g., if the approach was only tested on a few datasets or with a few runs. In general, empirical results often depend on implicit assumptions, which should be articulated.
        \item The authors should reflect on the factors that influence the performance of the approach. For example, a facial recognition algorithm may perform poorly when image resolution is low or images are taken in low lighting. Or a speech-to-text system might not be used reliably to provide closed captions for online lectures because it fails to handle technical jargon.
        \item The authors should discuss the computational efficiency of the proposed algorithms and how they scale with dataset size.
        \item If applicable, the authors should discuss possible limitations of their approach to address problems of privacy and fairness.
        \item While the authors might fear that complete honesty about limitations might be used by reviewers as grounds for rejection, a worse outcome might be that reviewers discover limitations that aren't acknowledged in the paper. The authors should use their best judgment and recognize that individual actions in favor of transparency play an important role in developing norms that preserve the integrity of the community. Reviewers will be specifically instructed to not penalize honesty concerning limitations.
    \end{itemize}

\item {\bf Theory assumptions and proofs}
    \item[] Question: For each theoretical result, does the paper provide the full set of assumptions and a complete (and correct) proof?
    \item[] Answer: \answerYes{} %
    \item[] Justification: The paper provides the full set of assumptions and a complete proof

    \item {\bf Experimental result reproducibility}
    \item[] Question: Does the paper fully disclose all the information needed to reproduce the main experimental results of the paper to the extent that it affects the main claims and/or conclusions of the paper (regardless of whether the code and data are provided or not)?
    \item[] Answer: \answerYes{} %
    \item[] Justification: The paper specifies the optimizer, training settings, and evaluation definitions in Appendices~\ref{app:y-recursion-derivation} and~\ref{app:multiseed}, together with the dataset and implementation details. Public reference implementations provide an avenue to reproduce the algorithm; the release is not an exhaustive archive of every historical run.
    \item[] Guidelines:
    \begin{itemize}
        \item The answer \answerNA{} means that the paper does not include experiments.
        \item If the paper includes experiments, a \answerNo{} answer to this question will not be perceived well by the reviewers: Making the paper reproducible is important, regardless of whether the code and data are provided or not.
        \item If the contribution is a dataset and\slash or model, the authors should describe the steps taken to make their results reproducible or verifiable. 
        \item Depending on the contribution, reproducibility can be accomplished in various ways. For example, if the contribution is a novel architecture, describing the architecture fully might suffice, or if the contribution is a specific model and empirical evaluation, it may be necessary to either make it possible for others to replicate the model with the same dataset, or provide access to the model. In general, releasing code and data is often one good way to accomplish this, but reproducibility can also be provided via detailed instructions for how to replicate the results, access to a hosted model (e.g., in the case of a large language model), releasing of a model checkpoint, or other means that are appropriate to the research performed.
        \item While NeurIPS does not require releasing code, the conference does require all submissions to provide some reasonable avenue for reproducibility, which may depend on the nature of the contribution. For example
        \begin{enumerate}
            \item If the contribution is primarily a new algorithm, the paper should make it clear how to reproduce that algorithm.
            \item If the contribution is primarily a new model architecture, the paper should describe the architecture clearly and fully.
            \item If the contribution is a new model (e.g., a large language model), then there should either be a way to access this model for reproducing the results or a way to reproduce the model (e.g., with an open-source dataset or instructions for how to construct the dataset).
            \item We recognize that reproducibility may be tricky in some cases, in which case authors are welcome to describe the particular way they provide for reproducibility. In the case of closed-source models, it may be that access to the model is limited in some way (e.g., to registered users), but it should be possible for other researchers to have some path to reproducing or verifying the results.
        \end{enumerate}
    \end{itemize}

\item {\bf Open access to data and code}
    \item[] Question: Does the paper provide open access to the data and code, with sufficient instructions to faithfully reproduce the main experimental results, as described in supplemental material?
    \item[] Answer: \answerNo{} %
    \item[] Justification: The implementation, example training configurations, and setup instructions are public at \url{https://github.com/agent-lab/hidden-ratio-adam}. This initial release provides pre-training, SFT, ReMax, and CIFAR examples, but does not provide a complete per-result manifest and all historical distribution-analysis scripts, so we do not claim fully scripted reproduction of every main result.

\item {\bf Experimental setting/details}
    \item[] Question: Does the paper specify all the training and test details (e.g., data splits, hyperparameters, how they were chosen, type of optimizer) necessary to understand the results?
    \item[] Answer: \answerYes{} %
    \item[] Justification: Mentioned in Appendix

\item {\bf Experiment statistical significance}
    \item[] Question: Does the paper report error bars suitably and correctly defined or other appropriate information about the statistical significance of the experiments?
    \item[] Answer: \answerNo{} %
    \item[] Justification: Appendix~\ref{app:multiseed} reports means and standard deviations over three seeds for selected pre-training and RLHF comparisons, and Appendix~\ref{app:architecture-scale} reports two-seed means and standard deviations for the CIFAR experiments. However, repeated-run uncertainty is not available for all main experiments or the additional diagnostic studies, so we retain a conservative No answer. These standard deviations describe variability across runs, not confidence intervals or significance tests.
    \item[] Guidelines:
    \begin{itemize}
        \item The answer \answerNA{} means that the paper does not include experiments.
        \item The authors should answer \answerYes{} if the results are accompanied by error bars, confidence intervals, or statistical significance tests, at least for the experiments that support the main claims of the paper.
        \item The factors of variability that the error bars are capturing should be clearly stated (for example, train/test split, initialization, random drawing of some parameter, or overall run with given experimental conditions).
        \item The method for calculating the error bars should be explained (closed form formula, call to a library function, bootstrap, etc.)
        \item The assumptions made should be given (e.g., Normally distributed errors).
        \item It should be clear whether the error bar is the standard deviation or the standard error of the mean.
        \item It is OK to report 1-sigma error bars, but one should state it. The authors should preferably report a 2-sigma error bar rather than state that they have a 96\% CI, if the hypothesis of Normality of errors is not verified.
        \item For asymmetric distributions, the authors should be careful not to show in tables or figures symmetric error bars that would yield results that are out of range (e.g., negative error rates).
        \item If error bars are reported in tables or plots, the authors should explain in the text how they were calculated and reference the corresponding figures or tables in the text.
    \end{itemize}

\item {\bf Experiments compute resources}
    \item[] Question: For each experiment, does the paper provide sufficient information on the computer resources (type of compute workers, memory, time of execution) needed to reproduce the experiments?
    \item[] Answer: \answerNo{} %
    \item[] Justification: The appendix identifies GPU types and counts for the original experiments and specifies training budgets. Per-run execution times, complete resource details for all additional studies, and total research compute are not available, so the disclosure is incomplete.
    
\item {\bf Code of ethics}
    \item[] Question: Does the research conducted in the paper conform, in every respect, with the NeurIPS Code of Ethics \url{https://neurips.cc/public/EthicsGuidelines}?
    \item[] Answer: \answerYes{} %
    \item[] Justification: The research conducted in the paper conforms, in every respect, with the NeurIPS Code of Ethics

\item {\bf Broader impacts}
    \item[] Question: Does the paper discuss both potential positive societal impacts and negative societal impacts of the work performed?
   \item[] Answer: \answerYes{}.
    \item[] Justification: The Broader Impact appendix discusses potential accessibility benefits of a future packed implementation and the possibility of lowering barriers to harmful downstream uses. It distinguishes these potential effects from measured results.

    
\item {\bf Safeguards}
    \item[] Question: Does the paper describe safeguards that have been put in place for responsible release of data or models that have a high risk for misuse (e.g., pre-trained language models, image generators, or scraped datasets)?
    \item[] Answer: \answerNA{}.
    \item[] Justification: The paper does not introduce or release a new high-risk pretrained model, image generator, or scraped dataset. The work focuses on optimizer-state representations and training methodology rather than releasing a deployable generative model.

\item {\bf Licenses for existing assets}
    \item[] Question: Are the creators or original owners of assets (e.g., code, data, models), used in the paper, properly credited and are the license and terms of use explicitly mentioned and properly respected?
    \item[] Answer:\answerNo{} %
    \item[] Justification: Original sources are cited and the code release preserves component license files. However, the paper does not provide a complete inventory of licenses and usage terms for every dataset, model, and software asset.

\item {\bf New assets}
    \item[] Question: Are new assets introduced in the paper well documented and is the documentation provided alongside the assets?
    \item[] Answer: \answerYes{}%
    \item[] Justification: We provide the source code and README documentation at \url{https://github.com/agent-lab/hidden-ratio-adam}.

\item {\bf Crowdsourcing and research with human subjects}
    \item[] Question: For crowdsourcing experiments and research with human subjects, does the paper include the full text of instructions given to participants and screenshots, if applicable, as well as details about compensation (if any)? 
    \item[] Answer: \answerNA{} %
    \item[] Justification: The paper does not involve crowdsourcing experiments or research with human subjects

\item {\bf Institutional review board (IRB) approvals or equivalent for research with human subjects}
    \item[] Question: Does the paper describe potential risks incurred by study participants, whether such risks were disclosed to the subjects, and whether Institutional Review Board (IRB) approvals (or an equivalent approval/review based on the requirements of your country or institution) were obtained?
    \item[] Answer: \answerNA{} %
    \item[] Justification: The paper does not involve crowdsourcing, user studies, or research with human subjects, so IRB approval or equivalent review is not applicable.


\item {\bf Declaration of LLM usage}
    \item[] Question: Does the paper describe the usage of LLMs if it is an important, original, or non-standard component of the core methods in this research? Note that if the LLM is used only for writing, editing, or formatting purposes and does \emph{not} impact the core methodology, scientific rigor, or originality of the research, declaration is not required.
    \item[] Answer: \answerNA{} %
    \item[] Justification: LLMs are not an important, original, or non-standard component of the proposed optimizer methodology.

\end{enumerate}

%% file: main_all.bbl
\begin{thebibliography}{41}
\providecommand{\natexlab}[1]{#1}
\providecommand{\url}[1]{\texttt{#1}}
\expandafter\ifx\csname urlstyle\endcsname\relax
  \providecommand{\doi}[1]{doi: #1}\else
  \providecommand{\doi}{doi: \begingroup \urlstyle{rm}\Url}\fi

\bibitem[Bai et~al.(2022)Bai, Jones, Ndousse, Askell, Chen, DasSarma, Drain, Fort, Ganguli, Henighan, Joseph, Kadavath, Kernion, Conerly, El-Showk, Elhage, Hatfield-Dodds, Hernandez, Hume, Johnston, Kravec, Lovitt, Nanda, Olsson, Amodei, Brown, Clark, McCandlish, Olah, Mann, and Kaplan]{bai2022training}
Yuntao Bai, Andy Jones, Kamal Ndousse, Amanda Askell, Anna Chen, Nova DasSarma, Dawn Drain, Stanislav Fort, Deep Ganguli, Tom Henighan, Nicholas Joseph, Saurav Kadavath, Jackson Kernion, Tom Conerly, Sheer El-Showk, Nelson Elhage, Zac Hatfield-Dodds, Danny Hernandez, Tristan Hume, Scott Johnston, Shauna Kravec, Liane Lovitt, Neel Nanda, Catherine Olsson, Dario Amodei, Tom Brown, Jack Clark, Sam McCandlish, Chris Olah, Ben Mann, and Jared Kaplan.
\newblock Training a helpful and harmless assistant with reinforcement learning from human feedback.
\newblock \emph{arXiv preprint arXiv:2204.05862}, 2022.

\bibitem[Balles and Hennig(2018)]{balles2018dissecting}
Lukas Balles and Philipp Hennig.
\newblock Dissecting adam: The sign, magnitude and variance of stochastic gradients.
\newblock In \emph{International Conference on Machine Learning}, 2018.

\bibitem[Bernstein et~al.(2018)Bernstein, Wang, Azizzadenesheli, and Anandkumar]{bernstein2018signsgd}
Jeremy Bernstein, Yu{-}Xiang Wang, Kamyar Azizzadenesheli, and Anima Anandkumar.
\newblock signsgd: Compressed optimisation for non-convex problems.
\newblock In \emph{International Conference on Machine Learning}, 2018.

\bibitem[Biderman et~al.(2023)Biderman, Schoelkopf, Anthony, Bradley, O'Brien, Hallahan, Khan, Purohit, Prashanth, Raff, Skowron, Sutawika, and van~der Wal]{biderman2023pythia}
Stella Biderman, Hailey Schoelkopf, Quentin Anthony, Herbie Bradley, Kyle O'Brien, Eric Hallahan, Mohammad~Aflah Khan, Shivanshu Purohit, USVSN~Sai Prashanth, Edward Raff, Aviya Skowron, Lintang Sutawika, and Oskar van~der Wal.
\newblock Pythia: A suite for analyzing large language models across training and scaling.
\newblock In \emph{International Conference on Machine Learning}, 2023.

\bibitem[Brown et~al.(2020)Brown, Mann, Ryder, Subbiah, Kaplan, Dhariwal, Neelakantan, Shyam, Sastry, Askell, Agarwal, Herbert-Voss, Krueger, Henighan, Child, Ramesh, Ziegler, Wu, Winter, Hesse, Chen, Sigler, Litwin, Gray, Chess, Clark, Berner, McCandlish, Radford, Sutskever, and Amodei]{brown2020language}
Tom~B. Brown, Benjamin Mann, Nick Ryder, Melanie Subbiah, Jared Kaplan, Prafulla Dhariwal, Arvind Neelakantan, Pranav Shyam, Girish Sastry, Amanda Askell, Sandhini Agarwal, Ariel Herbert-Voss, Gretchen Krueger, Tom Henighan, Rewon Child, Aditya Ramesh, Daniel~M. Ziegler, Jeffrey Wu, Clemens Winter, Christopher Hesse, Mark Chen, Eric Sigler, Mateusz Litwin, Scott Gray, Benjamin Chess, Jack Clark, Christopher Berner, Sam McCandlish, Alec Radford, Ilya Sutskever, and Dario Amodei.
\newblock Language models are few-shot learners.
\newblock In \emph{Advances in Neural Information Processing Systems}, 2020.

\bibitem[Cattaneo and Shigida(2026)]{cattaneo2026minibatchnoiseadam}
Matias~D. Cattaneo and Boris Shigida.
\newblock The effect of mini-batch noise on the implicit bias of adam.
\newblock \emph{arXiv preprint arXiv:2602.01642}, 2026.

\bibitem[Chen et~al.(2025)Chen, Hwang, and yi~Lee]{chen2025simpletransfer}
Kuang-Ming Chen, Jenq-Neng Hwang, and Hung yi~Lee.
\newblock {InstructionCP}: A simple yet effective approach for transferring large language models to target languages.
\newblock In \emph{Workshop on Research in Computational Linguistic Typology and Multilingual NLP}, 2025.

\bibitem[Chitsaz et~al.(2024)Chitsaz, Fournier, Mordido, and Chandar]{chitsaz2024exploring}
Kamran Chitsaz, Quentin Fournier, Gon{\c{c}}alo Mordido, and Sarath Chandar.
\newblock Exploring quantization for efficient pre-training of transformer language models.
\newblock In \emph{Findings of the Association for Computational Linguistics: EMNLP}, 2024.

\bibitem[Cui et~al.(2023)Cui, Yuan, Ding, Yao, He, Zhu, Ni, Xie, Xie, Lin, Liu, and Sun]{cui2023ultrafeedback}
Ganqu Cui, Lifan Yuan, Ning Ding, Guanming Yao, Bingxiang He, Wei Zhu, Yuan Ni, Guotong Xie, Ruobing Xie, Yankai Lin, Zhiyuan Liu, and Maosong Sun.
\newblock Ultrafeedback: Boosting language models with scaled ai feedback.
\newblock \emph{arXiv preprint arXiv:2310.01377}, 2023.

\bibitem[Devlin et~al.(2019)Devlin, Chang, Lee, and Toutanova]{devlin2019bert}
Jacob Devlin, Ming-Wei Chang, Kenton Lee, and Kristina Toutanova.
\newblock {BERT}: Pre-training of deep bidirectional transformers for language understanding.
\newblock In \emph{North American Chapter of the Association for Computational Linguistics}, 2019.

\bibitem[Ding et~al.(2023)Ding, Chen, Xu, Qin, Hu, Liu, Sun, and Zhou]{ding2023ultrachat}
Ning Ding, Yulin Chen, Bokai Xu, Yujia Qin, Shengding Hu, Zhiyuan Liu, Maosong Sun, and Bowen Zhou.
\newblock Enhancing chat language models by scaling high-quality instructional conversations.
\newblock In \emph{Conference on Empirical Methods in Natural Language Processing}, 2023.

\bibitem[Dohare et~al.(2023)Dohare, Lan, and Mahmood]{dohare2023policycollapse}
Shibhansh Dohare, Qingfeng Lan, and A.~Rupam Mahmood.
\newblock Overcoming policy collapse in deep reinforcement learning.
\newblock In \emph{Sixteenth European Workshop on Reinforcement Learning}, 2023.

\bibitem[Dohare et~al.(2024)Dohare, Hernandez-Garcia, Lan, Rahman, Mahmood, and Sutton]{lossplasticity2024}
Shibhansh Dohare, J.~Fernando Hernandez-Garcia, Qingfeng Lan, Parash Rahman, A.~Rupam Mahmood, and Richard~S. Sutton.
\newblock Loss of plasticity in deep continual learning.
\newblock \emph{Nature}, 2024.

\bibitem[Dosovitskiy et~al.(2021)Dosovitskiy, Beyer, Kolesnikov, Weissenborn, Zhai, Unterthiner, Dehghani, Minderer, Heigold, Gelly, Uszkoreit, and Houlsby]{dosovitskiy2021image}
Alexey Dosovitskiy, Lucas Beyer, Alexander Kolesnikov, Dirk Weissenborn, Xiaohua Zhai, Thomas Unterthiner, Mostafa Dehghani, Matthias Minderer, Georg Heigold, Sylvain Gelly, Jakob Uszkoreit, and Neil Houlsby.
\newblock An image is worth 16x16 words: Transformers for image recognition at scale.
\newblock In \emph{International Conference on Learning Representations}, 2021.

\bibitem[Ellis et~al.(2024)Ellis, Jackson, Lupu, Goldie, Fellows, Whiteson, and Foerster]{ellis2024adamrel}
Benjamin Ellis, Matthew~T. Jackson, Andrei Lupu, Alexander~D. Goldie, Mattie Fellows, Shimon Whiteson, and Jakob~N. Foerster.
\newblock Adam on local time: Addressing nonstationarity in rl with relative adam timesteps.
\newblock In \emph{Advances in Neural Information Processing Systems}, 2024.

\bibitem[Fern{\'a}ndez-Hern{\'a}ndez et~al.(2026)Fern{\'a}ndez-Hern{\'a}ndez, P{\'e}rez-Corral, Mestre, Dolz, and Quintana-Ort{\'i}]{fernandezhernandez2026adamscale}
Alberto Fern{\'a}ndez-Hern{\'a}ndez, Cristian P{\'e}rez-Corral, Jose~I. Mestre, Manuel~F. Dolz, and Enrique~S. Quintana-Ort{\'i}.
\newblock Why adam works better with {$\beta_1=\beta_2$}: The missing gradient scale invariance principle.
\newblock \emph{arXiv preprint arXiv:2601.21739}, 2026.

\bibitem[Fishman et~al.(2025)Fishman, Chmiel, Banner, and Soudry]{fishman2025scalingfp8}
Maxim Fishman, Brian Chmiel, Ron Banner, and Daniel Soudry.
\newblock Scaling fp8 training to trillion-token llms.
\newblock In \emph{International Conference on Learning Representations}, 2025.

\bibitem[Goldie et~al.(2024)Goldie, Lu, Jackson, Whiteson, and Foerster]{goldie2024learnedopt}
Alexander~D. Goldie, Chris Lu, Matthew~T. Jackson, Shimon Whiteson, and Jakob~N. Foerster.
\newblock Can learned optimization make reinforcement learning less difficult?
\newblock In \emph{Advances in Neural Information Processing Systems}, 2024.

\bibitem[Groeneveld et~al.(2024)Groeneveld, Beltagy, Walsh, Bhagia, Kinney, Tafjord, Jha, Ivison, Magnusson, Wang, et~al.]{groeneveld2024olmo}
Dirk Groeneveld, Iz~Beltagy, Pete Walsh, Akshita Bhagia, Rodney Kinney, Oyvind Tafjord, Ananya~Harsh Jha, Hamish Ivison, Ian Magnusson, Yizhong Wang, et~al.
\newblock {OLMo}: Accelerating the science of language models.
\newblock \emph{arXiv preprint arXiv:2402.00838}, 2024.

\bibitem[He et~al.(2019)He, Liu, and Tao]{he2019control}
Fengxiang He, Tongliang Liu, and Dacheng Tao.
\newblock Control batch size and learning rate to generalize well: Theoretical and empirical evidence.
\newblock In \emph{Advances in Neural Information Processing Systems}, volume~32, 2019.

\bibitem[He et~al.(2015)He, Zhang, Ren, and Sun]{he2015deepresidual}
Kaiming He, Xiangyu Zhang, Shaoqing Ren, and Jian Sun.
\newblock Deep residual learning for image recognition.
\newblock \emph{arXiv preprint arXiv:1512.03385}, 2015.

\bibitem[{Hugging FaceTB}(2025)]{smollm32025}
{Hugging FaceTB}.
\newblock {SmolLM3}: Smol, multilingual, long-context reasoner.
\newblock \url{https://huggingface.co/blog/smollm3}, 2025.

\bibitem[Kingma and Ba(2015)]{kingma2015adam}
Diederik~P. Kingma and Jimmy Ba.
\newblock Adam: A method for stochastic optimization.
\newblock In \emph{International Conference on Learning Representations}, 2015.

\bibitem[Krizhevsky(2009)]{krizhevsky2009learning}
Alex Krizhevsky.
\newblock Learning multiple layers of features from tiny images.
\newblock Technical report, University of Toronto, 2009.

\bibitem[Li et~al.(2022)Li, Wang, and Arora]{li2022zeroloss}
Zhiyuan Li, Tianhao Wang, and Sanjeev Arora.
\newblock What happens after {SGD} reaches zero loss? --a mathematical framework.
\newblock In \emph{International Conference on Learning Representations}, 2022.

\bibitem[Li et~al.(2024)Li, Xu, Zhang, Lin, Yu, Sun, and Luo]{li2024remax}
Ziniu Li, Tian Xu, Yushun Zhang, Zhihang Lin, Yang Yu, Ruoyu Sun, and Zhi{-}Quan Luo.
\newblock {ReMax}: A simple, effective, and efficient reinforcement learning method for aligning large language models.
\newblock In \emph{International Conference on Machine Learning}, 2024.

\bibitem[Liu et~al.(2025)Liu, Zeng, Xiao, He, Liu, Wang, Yan, Shen, Zhang, Xu, Liu, and Zhou]{liu2025skyworkrewardv2}
Chris~Yuhao Liu, Liang Zeng, Yuzhen Xiao, Jujie He, Jiacai Liu, Chaojie Wang, Rui Yan, Wei Shen, Fuxiang Zhang, Jiacheng Xu, Yang Liu, and Yahui Zhou.
\newblock Skywork-reward-v2: Scaling preference data curation via human-ai synergy.
\newblock \emph{arXiv preprint arXiv:2507.01352}, 2025.

\bibitem[Liu et~al.(2021)Liu, Lin, Cao, Hu, Wei, Zhang, Lin, and Guo]{liu2021swin}
Ze~Liu, Yutong Lin, Yue Cao, Han Hu, Yixuan Wei, Zheng Zhang, Stephen Lin, and Baining Guo.
\newblock Swin transformer: Hierarchical vision transformer using shifted windows.
\newblock In \emph{IEEE/CVF International Conference on Computer Vision}, 2021.

\bibitem[Liu et~al.(2020)Liu, Wang, and Ueda]{ziyin2020laprop}
Ziyin Liu, Zhikang~T. Wang, and Masahito Ueda.
\newblock Laprop: Separating momentum and adaptivity in adam.
\newblock \emph{arXiv preprint arXiv:2002.04839}, 2020.

\bibitem[Loshchilov and Hutter(2019)]{loshchilov2019decoupled}
Ilya Loshchilov and Frank Hutter.
\newblock Decoupled weight decay regularization.
\newblock In \emph{International Conference on Learning Representations}, 2019.

\bibitem[Mandt et~al.(2017)Mandt, Hoffman, and Blei]{mandt2017sgd}
Stephan Mandt, Matthew~D. Hoffman, and David~M. Blei.
\newblock Stochastic gradient descent as approximate bayesian inference.
\newblock \emph{Journal of Machine Learning Research}, 18\penalty0 (134):\penalty0 1--35, 2017.

\bibitem[Moalla et~al.(2024)Moalla, Miele, Pyatko, Pascanu, and Gulcehre]{moalla2024norepresentation}
Skander Moalla, Andrea Miele, Daniil Pyatko, Razvan Pascanu, and Caglar Gulcehre.
\newblock No representation, no trust: Connecting representation, collapse, and trust issues in ppo.
\newblock In \emph{Advances in Neural Information Processing Systems}, 2024.

\bibitem[Orvieto and Gower(2025)]{orvieto2025adamsecret}
Antonio Orvieto and Robert~M. Gower.
\newblock In search of adam's secret sauce.
\newblock In \emph{Advances in Neural Information Processing Systems}, 2025.

\bibitem[Ouyang et~al.(2022)Ouyang, Wu, Jiang, Almeida, Wainwright, Mishkin, Zhang, Agarwal, Slama, Ray, Schulman, Hilton, Kelton, Miller, Simens, Askell, Welinder, Christiano, Leike, and Lowe]{ouyang2022training}
Long Ouyang, Jeff Wu, Xu~Jiang, Diogo Almeida, Carroll~L. Wainwright, Pamela Mishkin, Chong Zhang, Sandhini Agarwal, Katarina Slama, Alex Ray, John Schulman, Jacob Hilton, Fraser Kelton, Luke Miller, Maddie Simens, Amanda Askell, Peter Welinder, Paul Christiano, Jan Leike, and Ryan Lowe.
\newblock Training language models to follow instructions with human feedback.
\newblock In \emph{Advances in Neural Information Processing Systems}, 2022.

\bibitem[Parmar et~al.(2024)Parmar, Satheesh, Patwary, Shoeybi, and Catanzaro]{parmar2024reuse}
Jupinder Parmar, Sanjev Satheesh, Mostofa Patwary, Mohammad Shoeybi, and Bryan Catanzaro.
\newblock Reuse, don't retrain: A recipe for continued pretraining of language models.
\newblock \emph{arXiv preprint arXiv:2407.07263}, 2024.

\bibitem[Penedo et~al.(2024)Penedo, Kydl{\'i}{\v{c}}ek, Allal, Lozhkov, Mitchell, Raffel, von Werra, and Wolf]{penedo2024fineweb}
Guilherme Penedo, Hynek Kydl{\'i}{\v{c}}ek, Loubna~Ben Allal, Anton Lozhkov, Margaret Mitchell, Colin Raffel, Leandro von Werra, and Thomas Wolf.
\newblock The fineweb datasets: Decanting the web for the finest text data at scale.
\newblock \emph{arXiv preprint arXiv:2406.17557}, 2024.

\bibitem[Peng et~al.(2023)Peng, Wu, Wei, Zhao, Yang, Liu, Xiong, Yang, Ni, Hu, Li, Zhang, Li, Ning, Wang, Zhang, Liu, Chau, Hu, and Cheng]{peng2023fp8lm}
Houwen Peng, Kan Wu, Yixuan Wei, Guoshuai Zhao, Yuxiang Yang, Ze~Liu, Yifan Xiong, Ziyue Yang, Bolin Ni, Jingcheng Hu, Ruihang Li, Miaosen Zhang, Chen Li, Jia Ning, Ruizhe Wang, Zheng Zhang, Shuguang Liu, Joe Chau, Han Hu, and Peng Cheng.
\newblock Fp8-lm: Training fp8 large language models.
\newblock \emph{arXiv preprint arXiv:2310.18313}, 2023.

\bibitem[Stiennon et~al.(2020)Stiennon, Ouyang, Wu, Ziegler, Lowe, Voss, Radford, Amodei, and Christiano]{stiennon2020learning}
Nisan Stiennon, Long Ouyang, Jeff Wu, Daniel~M. Ziegler, Ryan Lowe, Chelsea Voss, Alec Radford, Dario Amodei, and Paul Christiano.
\newblock Learning to summarize with human feedback.
\newblock In \emph{Advances in Neural Information Processing Systems}, 2020.

\bibitem[Su(2026)]{kexuefm2026betaeq}
Jianlin Su.
\newblock Are {$\beta_1=\beta_2$} the optimal hyperparameters for {Adam}?
\newblock \url{https://kexue.fm/archives/11593}, 2026.
\newblock Blog post in Chinese; title translated.

\bibitem[Sun et~al.(2023)Sun, Wang, Li, and Wang]{sun2023momentumsignsgd}
Tao Sun, Qingsong Wang, Dongsheng Li, and Bao Wang.
\newblock Momentum ensures convergence of signsgd under weaker assumptions.
\newblock In \emph{International Conference on Machine Learning}, 2023.

\bibitem[Touvron et~al.(2023)Touvron, Lavril, Izacard, Martinet, Lachaux, Lacroix, Rozi{\`e}re, Goyal, Hambro, Azhar, Rodriguez, Joulin, Grave, and Lample]{touvron2023llama}
Hugo Touvron, Thibaut Lavril, Gautier Izacard, Xavier Martinet, Marie-Anne Lachaux, Timoth{\'e}e Lacroix, Baptiste Rozi{\`e}re, Naman Goyal, Eric Hambro, Faisal Azhar, Aur{\'e}lien Rodriguez, Armand Joulin, Edouard Grave, and Guillaume Lample.
\newblock {LLaMA}: Open and efficient foundation language models.
\newblock \emph{arXiv preprint arXiv:2302.13971}, 2023.

\end{thebibliography}
